\documentclass[10pt,twocolumn,letterpaper]{article}

\usepackage[pagenumbers]{cvpr}              
\definecolor{cvprblue}{rgb}{0.21,0.49,0.74}
\usepackage[pagebackref,breaklinks,colorlinks,allcolors=cvprblue]{hyperref}
\usepackage[table]{xcolor}  
\usepackage{multirow}
\definecolor{tabfirst}{rgb}{1, 0.7, 0.7} 
\definecolor{tabsecond}{rgb}{1, 0.85, 0.7} 
\definecolor{tabthird}{rgb}{1, 1, 0.7} 
\usepackage{arydshln}
\usepackage{makecell}

\def\paperID{1374} 
\def\confName{CVPR}
\def\confYear{2026}

\title{Evaluating Low-Light Image Enhancement Across  Multiple Intensity Levels}

\author{Maria Pilligua$^{1,2}$ \quad David Serrano-Lozano$^{1,2}$ \quad Pai Peng$^{3}$ \\ Ramon Baldrich$^{1,2}$ \quad Michael S. Brown$^{4}$ \quad Javier Vazquez-Corral$^{1,2}$\\
{\normalsize $^1$Computer Vision Center} \quad
{\normalsize $^2$Universitat Autònoma de Barcelona} \\
{\normalsize $^3$University of Wisconsin-Madison} \quad
{\normalsize $^4$York University}\\
\small{\url{https://color.cvc.uab.cat/mill}}
}

\begin{document}

\maketitle


\begin{abstract}
Imaging in low-light environments is challenging due to reduced scene radiance, which leads to elevated sensor noise and reduced color saturation. Most learning-based low-light enhancement methods rely on paired training data captured under a single low-light condition and a well-lit reference. The lack of radiance diversity limits our understanding of how enhancement techniques perform across varying illumination intensities. We introduce the Multi-Illumination Low-Light (MILL) dataset, containing images captured at diverse light intensities under controlled conditions with fixed camera settings and precise illuminance measurements. MILL enables comprehensive evaluation of enhancement algorithms across variable lighting conditions. We benchmark several state-of-the-art methods and reveal significant performance variations across intensity levels. Leveraging the unique multi-illumination structure of our dataset, we propose improvements that enhance robustness across diverse illumination scenarios. Our modifications achieve up to 10 dB PSNR improvement for DSLR and 2 dB for the smartphone on Full HD images.
\end{abstract}

\vspace{-4mm}
\section{Introduction}

\begin{figure}
    \centering
    \includegraphics[width=\linewidth]{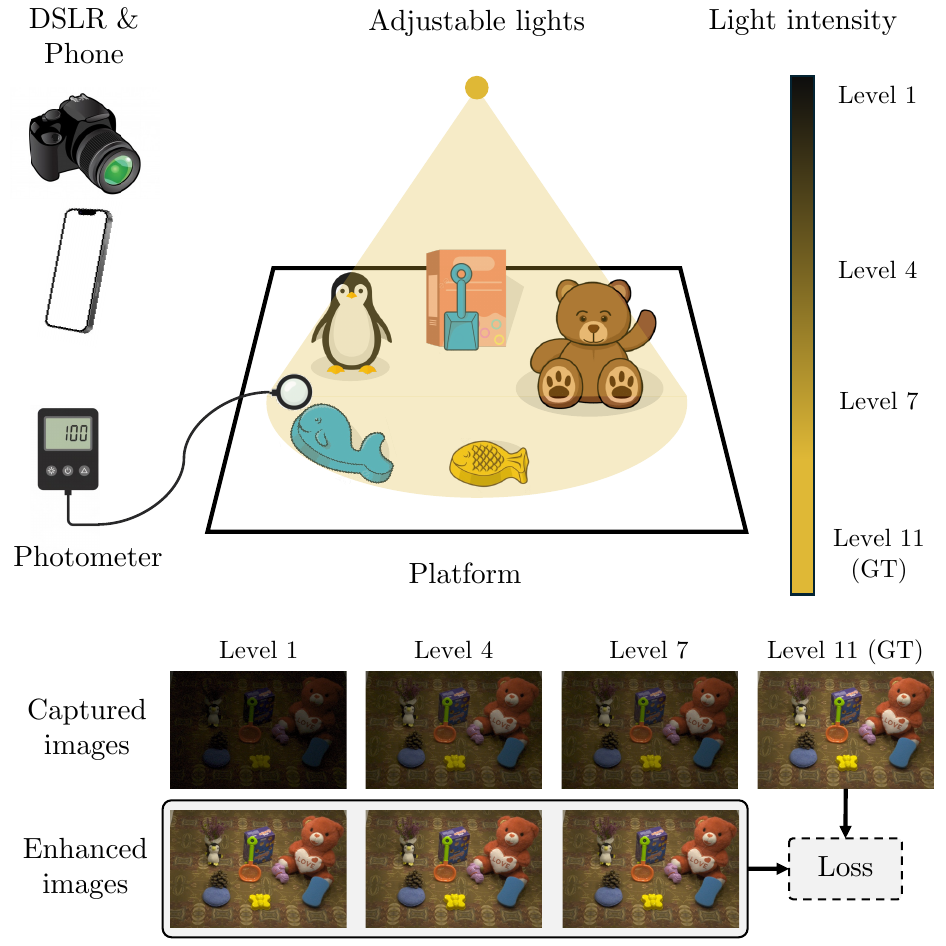}
    \caption{Illustration of our capture setup. A set of controllable lights illuminates the scene. For each scene, we capture 11 images by varying the light intensity from minimum to maximum at 10\% intervals (levels). Camera parameters (aperture, exposure time, ISO) remain fixed, and images are captured in unprocessed RAW format. The image captured at maximum intensity serves as the ground truth, while all other images serve as low-light inputs for training, validation, and testing. Scenes are captured with both a DSLR and a smartphone, and scene illuminance is measured with a photometer.}\label{fig:explanation}
    \vspace{-4mm}
    \label{fig:placeholder}
\end{figure}

Images taken in low-light environments are corrupted by sensor noise and diminished color saturation. Simple digital exposure adjustments, such as scaling the image's digital values, result in poor image quality due to high levels of sensor noise. Consequently, deep learning techniques have been developed to directly enhance low-light images, efficiently reducing noise and improving color and texture (e.g., \cite{RetinexNet, KinD, Ruas, Retinexformer, LLFormer}). The success of these approaches is heavily dependent on how the training data is collected.

Existing low-light image enhancement (LLIE) datasets obtain paired data either by varying camera settings or through post-processing, but nearly all capture images under a single low-light condition. This fails to reflect real-world scenarios where low-light images span a wide range of brightness levels, limiting the robustness of LLIE methods when deployed in practice.

To address this limitation, we present the Multi-Illumination Low-Light (MILL) dataset. Unlike existing datasets, MILL captures the same scene under 11 systematically varied light intensities, ranging from minimum to maximum brightness with equispaced intervals, while maintaining fixed camera parameters (see Fig.~\ref{fig:explanation}). Each capture is accompanied by precise illuminance measurements (lux) from a calibrated sensor and the input parameters of the programmable lights. We use the maximum-intensity image as ground truth and the remaining 10 images as low-light inputs. All images are captured in RAW format, ensuring no camera-processed artifacts.

Using the MILL dataset, we analyze how current state-of-the-art methods perform under varying lighting conditions and find that model performance varies significantly across different intensity ranges. Based on our findings, we propose an improvement over the best-performing method, Retinexformer~\cite{Retinexformer}. We propose to disentangle scene and illumination information in the network's latent features by leveraging the multi-level nature of our dataset. We demonstrate that our simple modification improves PSNR by $10$ dB for the DSLR and $2$ dB for the smartphone camera on Full HD images.  

\noindent Our contributions can be summarized as follows:
\begin{itemize}
\item We introduce {\bf MILL}, a new low-light image enhancement dataset in which each scene is captured at 11 distinct illumination levels. Every image is paired with a photometer-measured lux value and the corresponding input setting of the programmable lights. Using fixed camera parameters on both a DSLR and a smartphone, we collected a total of 1100 images.
\item We benchmark several state-of-the-art enhancement methods on our dataset to evaluate their robustness across a broad range of illumination levels. Our analysis reveals that certain methods display unexpected performance fluctuations at different intensity ranges.
\item We further propose two loss terms that exploit the auxiliary illumination information (i.e., intensity level) provided by our dataset. Integrating these terms leads to substantial improvements over prior state-of-the-art models.
\end{itemize}

\section{Related Work}
\subsection{Low-Light Datasets}

Early LLIE datasets, such as VV~\cite{VV} and LIME~\cite{LIME}, contained only unpaired low-light images (15 and 10 samples, respectively) without corresponding well-exposed references. For this reason, Wei et al.~\cite{Chen2018Retinex} introduced the LOw Light paired dataset (LOL) to allow end-to-end training of LLIE models. The LOL dataset proved valuable to the research community, enabling end-to-end training of methods. LoLv1 contains images captured under different camera settings to capture the same scene under low-light and well-lit conditions. The LOLv1 dataset consists of $500$ images, of which $485$ are for training and $15$ are for testing. An extension of the LoLv1 dataset, LoLv2, was later introduced~\cite{lolv2}. In LoLv2, the authors introduced two variants: one following the LoLv1 methodology and the other generating the low-light image synthetically from the well-lit counterpart. LoLv2 contains $689$ training scenes and $100$ test scenes. A major issue with these datasets is that they contain images from the same scene in both the training and test sets, potentially affecting generalization. 

Several datasets were subsequently introduced to address the limitations of early LLIE benchmarks, primarily in terms of scale and diversity. The DPED~\cite{DPED} dataset provided images captured across multiple smartphone cameras, enabling cross-device evaluation. Deep-UPE~\cite{DEEPUPE} emphasized extremely low-light scenarios with more challenging exposure conditions. The LSRW~\cite{LRSW} dataset expanded camera diversity by including both DSLR and smartphone captures, recognizing the distinct image formation characteristics of different sensor types. More recently, LLIV-Phone~\cite{LLIV-Phone} introduced temporal information by capturing video sequences under low-light conditions, allowing methods to exploit inter-frame correlations. This video-based approach was further explored in the DID~\cite{DID} and SDSD~\cite{SDSD} datasets, which provided paired low-light and normal-light video sequences for dynamic scene enhancement. 

A significant methodological shift came with the SID~\cite{SID} dataset, which captured image pairs in RAW format rather than processed RGB, enabling methods to leverage the complete sensor information before in-camera processing. To scale dataset creation, VE-LOL~\cite{VE-LOL} adopted a synthetic approach, darkening well-exposed images and adding synthetic noise patterns to simulate sensor characteristics at high ISO settings. Recently, the BVI-LowLight dataset~\cite{BVI-lowlight} was introduced, containing 40,000 images of objects captured at different ISO settings. Additionally, to obtain more training data, some LLIE methods use exposure-correction datasets such as PHOS~\cite{Phos} and SICE~\cite{SICE}. Despite these advances in scale, sensor diversity, temporal modeling, and data modality, existing datasets either capture each scene under a single low-light condition or rely on modifying the camera parameters. 



In this work, we introduce the Multi-Illumination Low-Light (MILL) dataset to address the limitations of existing low-light datasets. MILL is the first dataset to capture multiple low-light images of the same scene at varying illumination levels under fixed camera settings (i.e., constant ISO and shutter speed) by systematically controlling light intensity in a controlled environment. Each low-light image is paired with a corresponding ground truth captured under normal lighting. Additionally, we provide RAW files and accompanying metadata, including illumination intensity values and LUX measurements for each capture.

\begin{table}[t!]
    \centering
    \caption{Performance degradation of LLIE methods across varying brightness levels. Models trained on the original LoLv1 dataset show diminished performance when tested on blended images (20\% and 50\% ground truth mixing), despite reduced information loss. Lower $\Delta E_{76}$ and higher PSNR$_L$, the better.}
    \vspace{-2mm}
    \begin{tabular}{ccccc}
        \toprule 
         & \multicolumn{2}{c}{Retinexformer~\cite{Retinexformer}} & \multicolumn{2}{c}{CIDNet~\cite{hvi}}  \\
         \cmidrule{2-5}
         & $\Delta E_{76}$ & PSNR$_L$ & $\Delta E_{76}$ & PSNR$_L$\\
        \midrule
        Original & 8.810 & 28.819 & 10.587 & 26.381\\
        20\% & 11.450 & 21.910 & 16.981 & 17.721 \\
        50\% & 16.165 & 17.804 & 24.811 & 14.115\\
        \bottomrule
    \end{tabular}
    \vspace{-2mm}
    \label{tab:motivation}
\end{table}

\begin{figure}[t!] \centering
    \raisebox{.8\height}{\makebox[0.025\linewidth]{\rotatebox{90}{\makecell{\small Input}}}}
    \includegraphics[width=0.3\linewidth]{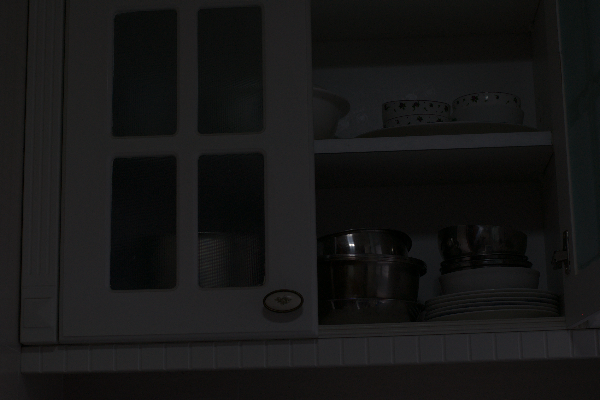}
    \includegraphics[width=0.3\linewidth]{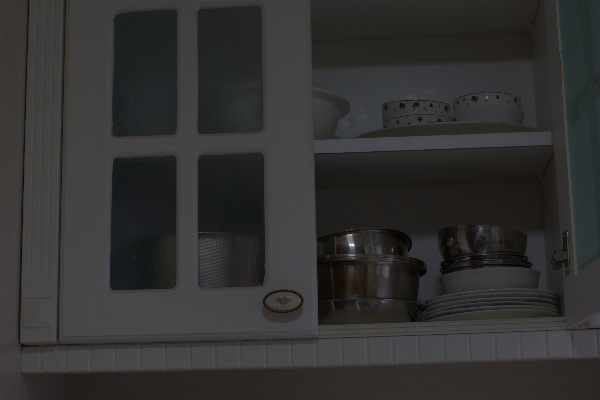}
    \includegraphics[width=0.3\linewidth]{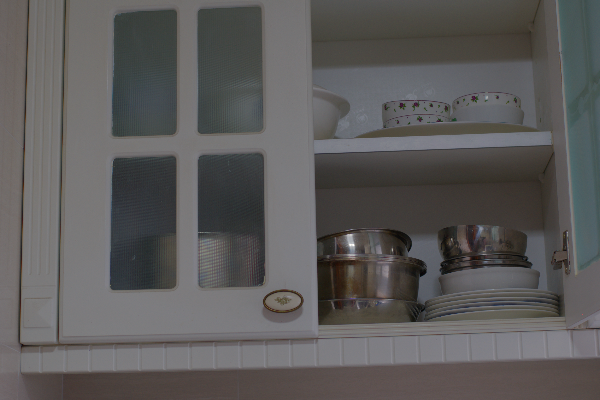}
    \\
    \raisebox{0.42\height}{\makebox[0.025\linewidth]{\rotatebox{90}{\makecell{\small Output}}}}
    \includegraphics[width=0.3\linewidth]{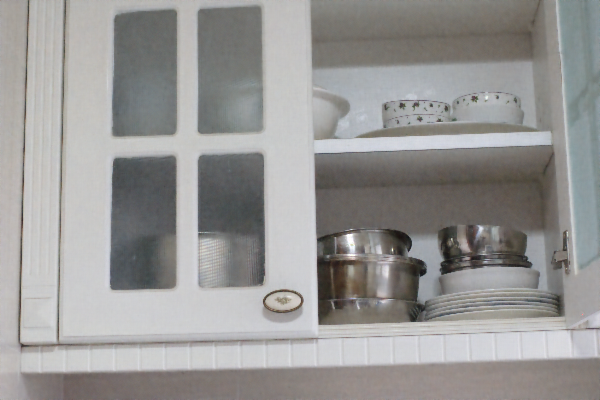}
    \includegraphics[width=0.3\linewidth]{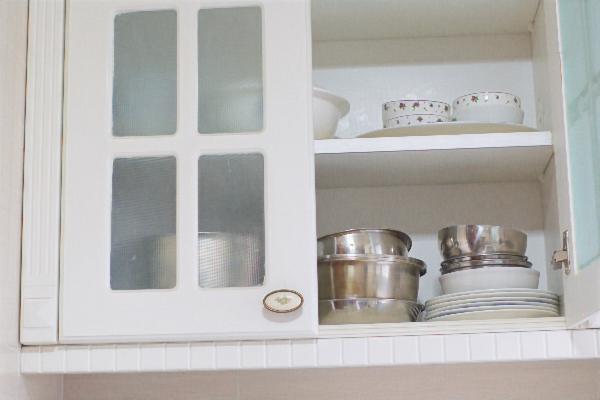}
    \includegraphics[width=0.3\linewidth]{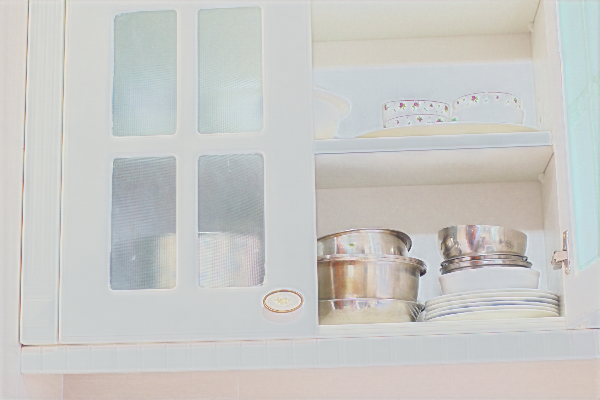}
    \\
    \makebox[0.025\linewidth]{}
    \makebox[0.3\linewidth]{\small Original}
    \makebox[0.3\linewidth]{\small 20\%}
    \makebox[0.3\linewidth]{\small 50\%}
    \vspace{-2mm}
    \caption{Impact of brightness variation on LLIE model performance. Blending input images with ground truth at 20\% and 50\% ratios degrades Retinexformer performance.}
    \vspace{-5mm}
    \label{fig:motivation}
\end{figure}

\subsection{Low-Light Image Enhancement Methods}

Early LLIE methods built upon the seminal Retinex algorithm~\cite{land1964retinex}, which decomposed images into reflectance and illumination components, inspiring variants including Multi-Scale Retinex~\cite{rahman1996multi}, SRIE~\cite{fu2016weighted}, and Milano-Retinex~\cite{milano1}. Methods such as LIME~\cite{LIME} and NPE~\cite{wang2013naturalness} demonstrated strong performance by leveraging natural image statistics without training data. However, these traditional approaches have been largely superseded by deep learning methods.

Early end-to-end deep learning methods include SID~\cite{SID} for RAW images and RetinexNet~\cite{Chen2018Retinex} for RGB inputs. Subsequent approaches introduced various architectural innovations: GLADNet~\cite{wang2018gladnet} combined global illumination and local detail modules; KinD~\cite{KinD} and KinD++~\cite{zhang2021beyond} adopted Retinex-inspired decomposition strategies; and Yang et al.~\cite{yang2020fidelity} incorporated adversarial learning. Recent methods leverage transformer architectures (LLFormer~\cite{LLFormer, LLFormer2}, Retinexformer~\cite{Retinexformer}) and diffusion models (Diff-Retinex~\cite{yi2023diff}, PyDiff~\cite{pydiff}). Alternative formulations include specialized color spaces~\cite{hvi} and pixel-wise mean estimation losses~\cite{liao2025gt}. Some methods jointly address enhancement and degradation removal, such as DarkIR~\cite{darkir} and LEDNet~\cite{zhou2022lednet}. To avoid paired training data requirements, unsupervised approaches have been proposed, including EnlightenGAN~\cite{jiang2021enlightengan}, Zero-DCE~\cite{guo2020zero}, SCI~\cite{SCI}, and lightweight RUAS~\cite{Ruas}.

However, all existing methods have been evaluated exclusively on fixed single low-light inputs without considering behavior across varying illumination levels. We benchmark state-of-the-art LLIE methods across different brightness conditions and propose two simple modifications to Retinexformer~\cite{Retinexformer} that leverage our multi-illumination dataset to improve robustness across intensity levels.

\section{Multi-Illumination Low-Light (MILL) Dataset}\label{sec:dataset}

\begin{figure*}[t!]
    \centering
    \includegraphics[width=0.16\textwidth]{new_figures/dataset_images/Scene2_a-D-B_0.png}
    \includegraphics[width=0.16\textwidth]{new_figures/dataset_images/Scene2_a-D-B_80.png}
    \includegraphics[width=0.16\textwidth]{new_figures/dataset_images/Scene2_a-D-B_113.png}
    \includegraphics[width=0.16\textwidth]{new_figures/dataset_images/Scene2_a-D-B_180.png}
    \includegraphics[width=0.16\textwidth]{new_figures/dataset_images/Scene2_a-D-B_213.png}
    \includegraphics[width=0.16\textwidth]{new_figures/dataset_images/Scene2_a-D-B_254.png}\\

    \includegraphics[width=0.16\textwidth]{new_figures/dataset_images/Val7-D-B_0.png}
    \includegraphics[width=0.16\textwidth]{new_figures/dataset_images/Val7-D-B_80.png}
    \includegraphics[width=0.16\textwidth]{new_figures/dataset_images/Val7-D-B_113.png}
    \includegraphics[width=0.16\textwidth]{new_figures/dataset_images/Val7-D-B_180.png}
    \includegraphics[width=0.16\textwidth]{new_figures/dataset_images/Val7-D-B_213.png}
    \includegraphics[width=0.16\textwidth]{new_figures/dataset_images/Val7-D-B_254.png}
    \\

    \includegraphics[width=0.16\textwidth]{new_figures/dataset_images/Test14-S-B_0.png}
    \includegraphics[width=0.16\textwidth]{new_figures/dataset_images/Test14-S-B_80.png}
    \includegraphics[width=0.16\textwidth]{new_figures/dataset_images/Test14-S-B_113.png}
    \includegraphics[width=0.16\textwidth]{new_figures/dataset_images/Test14-S-B_180.png}
    \includegraphics[width=0.16\textwidth]{new_figures/dataset_images/Test14-S-B_213.png}
    \includegraphics[width=0.16\textwidth]{new_figures/dataset_images/Test14-S-B_254.png}
    \\
    \makebox[0.16\linewidth]{\small Level 1}
    \makebox[0.16\linewidth]{\small Level 2}
    \makebox[0.16\linewidth]{\small Level 3}
    \makebox[0.16\linewidth]{\small Level 6}
    \makebox[0.16\linewidth]{\small Level 8}
    \makebox[0.16\linewidth]{\small Ground Truth}
    \vspace{-2mm}
    \caption{Example scenes from our dataset at different levels for both the DSLR (first two rows) and the smartphone camera (last row).}
    \vspace{-4mm}

    \label{fig:dataset}
\end{figure*}

Existing LLIE datasets present two critical limitations: (1) they either contain a single severely underexposed image per scene (e.g.~\cite{lolv2, Chen2018Retinex}) or (2) simulate brightness variations via camera parameter adjustments or post-processing (e.g.~\cite{BVI-lowlight, SICE, Phos}). This constraint limits real-world applicability, where low-light conditions span a continuous range of intensities.

To quantify this limitation, we simulated varying brightness levels on the LoLv1 dataset~\cite{RetinexNet} by blending input images with their ground truth counterparts at different ratios. This blending reduces degradation severity by simulating intermediate brightness levels, theoretically making enhancement easier. As shown in Table~\ref{tab:motivation}, both Retinexformer~\cite{Retinexformer} and CIDNet~\cite{hvi} performed worse on the blended versions (with 0.2 and 0.5 ground truth mixing ratios) than on the original dataset, as measured by $\Delta E_{76}$ and PSNR$_L$. This result occurs because models trained on fixed brightness levels fail to generalize across different intensities. Figure~\ref{fig:motivation} illustrates this problem. Oversaturation in the output increases proportionally with input brightness, a clear evidence that intermediate brightness levels are absent from training data. This lack of intensity diversity severely limits the practical applicability of LLIE methods, as real-world deployment requires robustness across varying brightness conditions.
To study this problem and address this limitation, we introduce a novel LLIE dataset featuring multiple brightness levels per scene with fixed camera parameters, enabling more robust training, benchmarking, and evaluation of LLIE methods across multiple intensity levels.

Our primary objective is to capture a high-quality, well-calibrated dataset for evaluating and training LLIE methods across different intensity levels. To ensure consistency and eliminate uncontrollable variables, we captured all images in a controlled indoor environment without windows or external light sources. We used a dedicated room with a platform for placing different floor backgrounds and objects, equipped with programmable lighting to precisely control brightness levels. Images were captured using two devices: a Nikon D5200 DSLR camera and a Samsung Galaxy S7 smartphone. The smartphone provides a contrasting capture profile compared to the DSLR, enabling evaluation across different sensor characteristics. We employed nine Philips Hue lights, whose emission spectrum covers the full visible range and provides sufficient spectral support for our experiments. Although we acknowledge that LED-based illuminants can exhibit spectral peaks and may not perfectly reproduce the continuous spectra of natural Planckian light sources, they are the preferred technology for fine-grained intensity control. Figure~\ref{fig:explanation} provides a schematic overview.

During image capture, all camera parameters remained fixed while scene light intensity was adjusted to achieve the desired brightness level. To capture well-lit ground-truth images, we set programmable light sources to maximum power without oversaturating the scene. We fixed the ISO to 100 and placed a Macbeth color chart at the platform center. For the DSLR, we systematically tested all aperture-shutter speed combinations, while for the smartphone (fixed aperture), we tested all available shutter speeds. We analyzed the RGB values of the white patch in the color chart and selected the image with values closest to 95\% of the maximum intensity in the camera-RAW format before saturation. This process determined optimal settings of f/9 and 1/5 seconds for the DSLR, and f/1.7 (default) and 1/10 seconds for the smartphone. 

We captured the lowest-intensity images (Level 1) with lights at minimum power. To obtain intermediate brightness levels, we computed 10 evenly spaced intervals based on lux meter readings between Level 1 and the ground truth, creating Levels 2-10, where lower numbers correspond to lower illumination intensity.

\begin{table*}[t!]
\caption{Performance of different LLIE methods across different intensity levels on our DSLR split of the MILL-s dataset. We report the mean $\Delta E_{76}$ and PSNR on the luminance channel (PSNR$_L$). \colorbox{tabfirst}{Best}, \colorbox{tabsecond}{second best}, and \colorbox{tabthird}{third best} results are highlighted.}
\vspace{-2mm}
\small
\setlength{\tabcolsep}{3.5pt}
\begin{tabular}{lclcccccccccc}
\toprule
& & \multicolumn{1}{c}{} & \multicolumn{2}{c}{Level 1}           & \multicolumn{2}{c}{Level 3}          & \multicolumn{2}{c}{Level 5}          & \multicolumn{2}{c}{Level 7}          & \multicolumn{2}{c}{Level 9}          \\ \cmidrule{4-13} 
& \multicolumn{1}{c}{Params (M)} & \multicolumn{1}{c}{Venue} & \multicolumn{1}{l}{$\Delta E_{76}$} & PSNR$_L$ & \multicolumn{1}{l}{$\Delta E_{76}$} & PSNR$_L$ & \multicolumn{1}{l}{$\Delta E_{76}$} & PSNR$_L$ & \multicolumn{1}{l}{$\Delta E_{76}$} & PSNR$_L$ & \multicolumn{1}{l}{$\Delta E_{76}$} & PSNR$_L$ \\ \midrule
              
Unprocessed & \multicolumn{1}{c}{-} & \multicolumn{1}{c}{-} & 30.34 & 13.46 & 18.62 & 17.68 & 11.90 & 21.48 & 7.60 & 25.56 & 3.62 & 36.64 \\

\hdashline

RUAS \cite{Ruas} & 0.003 & CVPR'21 & 25.46 & 16.63 & 45.33 & 9.86 & 57.47 & 6.93 & 62.99 & 5.86 & 67.14 & 5.18 \\ 
LLFormer \cite{LLFormer2} & 24.52 & AAAI'23 & 16.37 & 20.88 & 13.73 & 21.79 & 13.34 & 22.06 & 13.17 & 22.25 & 12.90 & 22.55 \\ 
KinD \cite{KinD} & 1.20 & ACMM'19 & 23.88 & 16.71 & 17.62 & 21.79 & 14.87 & 21.34 & 14.49 & 21.63 & 15.00 & 21.36 \\ 
FourLLIE \cite{wang2023fourllie} & 0.12 & ACMM'23 & 24.51 & 17.29 & 22.79 & 17.66 & 26.64 & 14.97 & 28.79 & 14.07 & 30.79 & 13.35 \\ 
SCI  \cite{SCI} & 0.0003 & CVPR'22 & 24.05 & 16.02 & 17.99 & 21.42 & 25.89 & 15.69 & 31.66 & 13.24 & 38.38 & 11.23 \\ 
MirNet \cite{mirnet} & 5.86 & CVPR'20 & \cellcolor{tabthird}14.03 & \cellcolor{tabfirst}26.46 & 11.11 & 25.34 & 11.39 & 24.81 & 11.65 & 24.49 & 11.72 & 24.96 \\ 
Retinexformer \cite{Retinexformer} & 1.61 & ICCV'23 & 14.15 & 25.09 & \cellcolor{tabsecond}10.45 & \cellcolor{tabsecond}26.39 & \cellcolor{tabsecond}10.35 & \cellcolor{tabsecond}26.55 & \cellcolor{tabsecond}10.41 & \cellcolor{tabsecond}26.48 & \cellcolor{tabthird}10.46 & \cellcolor{tabsecond}27.41 \\
DarkIR~\cite{darkir} & 3.31 & CVPR'25 & 14.39 & 24.65 & 11.29 & 25.23 & 11.58 & 24.74 & \cellcolor{tabsecond}10.41 & 23.91 & 12.15 & 24.63 \\
HVI-CIDNet~\cite{hvi} & 1.88 & CVPR'25 & 14.78 & 24.08 & 13.71 & 22.49 & 14.58 & 21.44 & 15.22 & 20.83 & 15.85 & 20.63 \\
PromptNorm~\cite{promptnorm} & 44.80 & CVPRW'25 & \cellcolor{tabfirst}13.47 & \cellcolor{tabsecond}25.89 & \cellcolor{tabthird}10.51 & \cellcolor{tabthird}26.06 & \cellcolor{tabthird}10.59 & \cellcolor{tabthird}25.94 & 10.82 & \cellcolor{tabthird}25.66 & \cellcolor{tabsecond}10.89 & \cellcolor{tabthird}26.28 \\
GT-Mean~\cite{liao2025gt} & 1.88 & ICCV'25 & 14.59 & 24.32 & 12.48 & 23.76 & 13.23 & 22.80 & 13.80 & 22.19 & 13.57 & 22.88 \\
Ours & 1.61 & \multicolumn{1}{c}{-} & \cellcolor{tabsecond}13.90 & \cellcolor{tabthird}25.53 & \cellcolor{tabfirst}9.11 & \cellcolor{tabfirst}31.47 & \cellcolor{tabfirst}9.09 & \cellcolor{tabfirst}31.52 & \cellcolor{tabfirst}8.94 & \cellcolor{tabfirst}32.31 & \cellcolor{tabfirst}9.17 & \cellcolor{tabfirst}32.48 \\
\hline
\end{tabular}
\vspace{-4mm}
\label{tab:levels}
\end{table*}

We assembled 6 different backgrounds and 98 different objects, with no overlap between train/validation and test sets. The dataset comprises 4 backgrounds in training/validation scenes and 2 in test scenes, with 46 unique objects for training, 24 for validation, and 28 for testing. We captured 50 scenes using both the DSLR and the smartphone across all 11 intensity levels, totaling 1,100 images. The dataset is split into 30 training, 12 validation, and 8 test scenes. Figure~\ref{fig:dataset} shows three representative scenes displaying some of the intensity levels to illustrate the illumination intervals. We display the three lowest values to show how the intensity levels change in consecutive levels. Note that the highest levels remain noticeably underexposed compared to the ground truth, demonstrating the continuous range of realistic low-light conditions.

All images were captured in RAW format (NEF for DSLR, DNG for smartphone) and processed using Camera RAW. Our RAW-to-sRGB processing was deliberately kept as minimally invasive as possible, since low-light enhancement is expected to operate at the early stages of the camera pipeline. DSLR images have a native resolution of 6036$\times$4020 pixels, while smartphone images are 1560$\times$1040 pixels. Following prior LLIE datasets, we created a small version (MILL-s) by bilinearly resizing all images to 600$\times$400 pixels to enable evaluation of methods with computational or memory constraints. Additionally, we divided each DSLR image into 9 non-overlapping patches of 2012$\times$1340 pixels, expanding the dataset to 5,500 Full-HD resolution images. Smartphone images remained at their original resolution due to their comparable full-HD size. We refer to this higher-resolution variant as MILL-f.

\section{Method using New Loss Terms}

We introduce two auxiliary loss terms that leverage the multi-level nature of our dataset to improve existing LLIE methods. Our goal is to explicitly disentangle the latent features into illumination-related and scene-related components. To this end, we introduce two complementary constraints: (1) an intensity prediction loss that uses the first latent channel to predict the input illumination level, and (2) a scene consistency loss that encourages the remaining channels to encode illumination-invariant scene content across different brightness conditions. The following subsections describe each loss term in detail before presenting the full objective.

Most current LLIE architectures follow a UNet-like structure, comprising an encoder and a decoder. We aim to disentangle the latent features extracted by the architecture's bottleneck. We adopt Retinexformer~\cite{Retinexformer} as our baseline architecture due to its strong performance in our benchmark evaluation (see Section~\ref{sec:benchmark}).

\subsection{Intensity Prediction Loss}
We propose a straightforward approach to encode the scene intensity level using the latent features. Specifically, we constrain the first latent feature channel to predict the normalized intensity value of the scene, $i_{in} \in [0,1]$.

To accomplish this, we introduce a loss component, $\mathcal{L}_{ip}$, that minimizes the $L_1$ distance between the predicted intensity at each spatial location and the known scene illumination intensity. Let $\mathbf{Z}_I \in \mathbb{R}^{H \times W}$ denote the first channel of the latent features and $I_{in} \in \mathbb{R}^{H \times W}$ denote the spatially-replicated version of $i_{in}$ matching the spatial dimensions of $\mathbf{Z}_I$. The intensity prediction loss is defined as:
\begin{equation}
    \mathcal{L}_{i} = ||\mathbf{Z}_I - I_{in}||_{1}.
\end{equation}

\begin{table*}[t!] 
\caption{Quantitative comparisons on our MILL-s for the DSLR and the smartphone splits. Results are averaged over all the images.  \colorbox{tabfirst}{Best}, \colorbox{tabsecond}{second best}, and \colorbox{tabthird}{third best} results are highlighted.}
\vspace{-2mm}
\centering
\small
\begin{tabular}{llcccccccc}
\toprule
& & PSNR$_L$ $\uparrow$ & PSNR$_{C}$  $\uparrow$ & SSIM $\uparrow$ & LPIPS $\downarrow$ & $\Delta$E$_{76}$ $\downarrow$ & MS-SWD $\downarrow$ & NIQE $\downarrow$ & Brisque $\downarrow$ \\ \midrule
\multirow{15}{*}{\rotatebox{90}{DSLR}} & Unprocessed & 23.237 & 21.718 & 0.740 & 0.151 & 13.211 & 1.615 & 5.419 & 18.226 \\
\hdashline
& RUAS~\cite{Ruas} & 8.310 & 6.497 & 0.355 & 0.499 & 53.752 & 6.781 & 6.202 & 29.091 \\
& LLFormer~\cite{LLFormer} & 22.029 & 19.731 & 0.850 & 0.155 & 13.622 & 1.490 & 3.876 & \cellcolor{tabsecond}10.207 \\
& KinD~\cite{KinD} & 20.409 & 18.157 & 0.779 & 0.219 & 16.782 & 1.871 & 4.383 & 15.948 \\
& FourLLIE~\cite{wang2023fourllie} & 15.486 & 13.534 & 0.687 & 0.241 & 26.759 & 3.514 & 5.398 & 19.201 \\
& SCI~\cite{SCI} & 15.329 & 13.006 & 0.609 & 0.270 & 28.669 & 3.407 & 5.791 & 26.788 \\
& MirNet~\cite{mirnet} & 25.141 & 21.821 & 0.882 & \cellcolor{tabsecond}0.139 & 11.763 & 1.403 & 3.953 & 14.382 \\
& Retinexformer~\cite{Retinexformer} & \cellcolor{tabsecond}26.557 & \cellcolor{tabsecond}22.782 & \cellcolor{tabthird}0.888 & \cellcolor{tabthird}0.140 & \cellcolor{tabsecond}10.944 & \cellcolor{tabsecond}1.254 & 3.880 & 12.197 \\
& MambaLLIE~\cite{weng2024mamballie} & 25.989 & \cellcolor{tabthird}22.518 & 0.886 & 0.141 & 11.984 & 1.387 & 3.781 & 13.861\\
& CWNet~\cite{zhang2025cwnet} & 25.490 & 22.231 & 0.883 & 0.146 & 11.722 & 1.492 & 3.992 & 14.853 \\
& DarkIR~\cite{darkir} & 24.700 & 21.502 & 0.876 & 0.142 & 12.177 & 1.382 & 3.788 & 12.351 \\
& HVI-CIDNet~\cite{hvi} & 21.755 & 19.173 & 0.844 & 0.155 & 14.873 & 1.699 & \cellcolor{tabfirst}3.727 & 12.160 \\
& PromptNorm~\cite{promptnorm} & \cellcolor{tabthird}26.061 & 22.497 & \cellcolor{tabsecond}0.893 & 0.144 & \cellcolor{tabthird}11.059 & \cellcolor{tabthird}1.308 & 3.779 & \cellcolor{tabthird}11.646 \\
& GT-Mean~\cite{liao2025gt} & 23.083 & 20.257 & 0.861 & 0.147 & 13.520 & 1.539 & \cellcolor{tabsecond}3.744 & 11.817 \\
& Ours & \cellcolor{tabfirst}31.209 & \cellcolor{tabfirst}26.197 & \cellcolor{tabfirst}0.896 & \cellcolor{tabfirst}0.135 & \cellcolor{tabfirst}9.681 & \cellcolor{tabfirst}1.036 & \cellcolor{tabthird}3.754 & \cellcolor{tabfirst}9.550 \\
\midrule
\multirow{15}{*}{\rotatebox{90}{Smartphone}} & Unprocessed & 19.155 & 17.573 & 0.511 & 0.215 & 17.832 & 2.718 & 5.088 & 20.895 \\
\hdashline
 & RUAS~\cite{Ruas} & 6.920 & 5.082 & 0.267 & 0.722 & 62.707 & 6.246 & 10.556 & 71.760 \\
&LLFormer~\cite{LLFormer} & \cellcolor{tabthird}23.015 & \cellcolor{tabthird}20.062 & 0.580 & 0.203 & \cellcolor{tabthird}12.510 & \cellcolor{tabfirst}1.518 & 4.194 & 20.877 \\
& KinD~\cite{KinD} & 19.462 & 17.452 & 0.536 & 0.259 & 17.252 & 1.980 & 3.546 & 22.950 \\
& FourLLIE~\cite{wang2023fourllie} & 21.377 & 18.510 & 0.539 & 0.236 & 16.976 & 3.279 & 4.831 & 22.159 \\
&SCI~\cite{SCI} & 15.960 & 13.208 & 0.437 & 0.330 & 29.477 & 3.314 & 5.475 & 24.372 \\
&MirNet~\cite{mirnet} & 20.760 & 18.100 & 0.614 & 0.219 & 15.231 & 1.720 & 3.537 & 23.269 \\
&Retinexformer~\cite{Retinexformer} & \cellcolor{tabsecond}23.230 & \cellcolor{tabsecond}20.485 & \cellcolor{tabsecond}0.629 & \cellcolor{tabthird}0.195 & \cellcolor{tabsecond}12.162 & \cellcolor{tabthird}1.682 & 3.162 & 20.821 \\
& MambaLLIE~\cite{weng2024mamballie} & 22.638 & 18.763 & 0.608 & 0.214 & 13.429 & 2.193 & 3.465 & 20.463 \\
& CWNet~\cite{zhang2025cwnet} & 22.432 & 18.532 & 0.601 & 0.219 & 13.945 & 2.374 & 3.469 & 20.642 \\
&DarkIR~\cite{darkir} & 22.540 & 19.898 & 0.622 & \cellcolor{tabfirst}0.192 & 13.047 & 1.679 & \cellcolor{tabsecond}3.070 & 19.388 \\
&HVI-CIDNet~\cite{hvi} & 20.517 & 18.170 & 0.598 & 0.196 & 15.823 & 1.782 & \cellcolor{tabthird}3.080 & \cellcolor{tabfirst}17.016 \\
&PromptNorm~\cite{promptnorm} & 22.198 & 19.464 & \cellcolor{tabthird}0.627 & 0.206 & 13.238 & 1.692 & 3.477 & 22.612 \\
& GT-Mean~\cite{liao2025gt} & 21.503 & 19.018 & 0.610 & \cellcolor{tabfirst}0.192 & 14.438 & 1.716 & \cellcolor{tabfirst}3.037 & \cellcolor{tabsecond}17.877 \\
&Ours & \cellcolor{tabfirst}23.870 & \cellcolor{tabfirst}21.166 & \cellcolor{tabfirst}0.629 & \cellcolor{tabthird}0.195 & \cellcolor{tabfirst}11.671 & \cellcolor{tabsecond}1.619 & 3.235 & \cellcolor{tabthird}18.733 \\
\bottomrule
\vspace{-7mm}
\end{tabular}
\label{tab:avg}
\end{table*}

\subsection{Scene Content Loss}
While the intensity prediction loss constrains the first channel to encode illumination information, we enforce the remaining channels to focus on scene content independent of lighting conditions. We achieve this through a triplet loss that encourages images of the same scene captured under different illumination levels to have similar latent representations (excluding the intensity channel), while pushing apart representations of different scenes captured at the same intensity level.

The scene content loss $\mathcal{L}_{s}$ is defined as:
\begin{equation}
 \mathcal{L}_{s} = \max\left(||\mathbf{Z}_q - \mathbf{Z}_p||^2 + m - ||\mathbf{Z}_q - \mathbf{Z}_n||^2, 0\right),
\end{equation}
where $\mathbf{Z}_q, \mathbf{Z}_p, \mathbf{Z}_n \in \mathbb{R}^{H \times W \times (C-1)}$ correspond to the latent features (excluding the intensity channel) of three images: the query input image, a positive sample from the same scene with different illumination, and a negative sample from a different scene with the same brightness level as the query. The margin $m$ defines the minimum desired distance between positive and negative pairs; we set $m = 1$ in all experiments.

\subsection{Combined Objective Function}
In addition to the proposed loss terms, we employ a reconstruction loss, $\mathcal{L}_{re}$, defined as the $L_1$ distance between the network output and the ground truth image. The complete objective function combines all three components:
\begin{equation}
\mathcal{L} = \mathcal{L}_{re} + \mathcal{L}_{i} + \mathcal{L}_{s}.
\end{equation}

\section{Experiments}

\subsection{Benchmark on MILL}\label{sec:benchmark}
We benchmark mainstream LLIE methods by retraining them on our MILL-s dataset using their officially released code. Our evaluation includes unsupervised methods (RUAS~\cite{Ruas}, SCI~\cite{SCI}), a Retinex-based approach (KinD~\cite{KinD}), transformer-based methods built on Restormer~\cite{restormer}~(LLFormer~\cite{LLFormer}, Retinexformer~\cite{Retinexformer}, PromptNorm~\cite{promptnorm}), a frequency-domain method (FourLLIE~\cite{wang2023fourllie}), image restoration approaches (MIRNet~\cite{mirnet}, DarkIR~\cite{darkir}), and specialized LLIE methods (HVI-CIDNet~\cite{hvi}, GT-Mean~\cite{liao2025gt}). We also evaluate our proposed modifications to Retinexformer.

\begin{figure*}[t!]
    \centering
    \includegraphics[width=\textwidth]{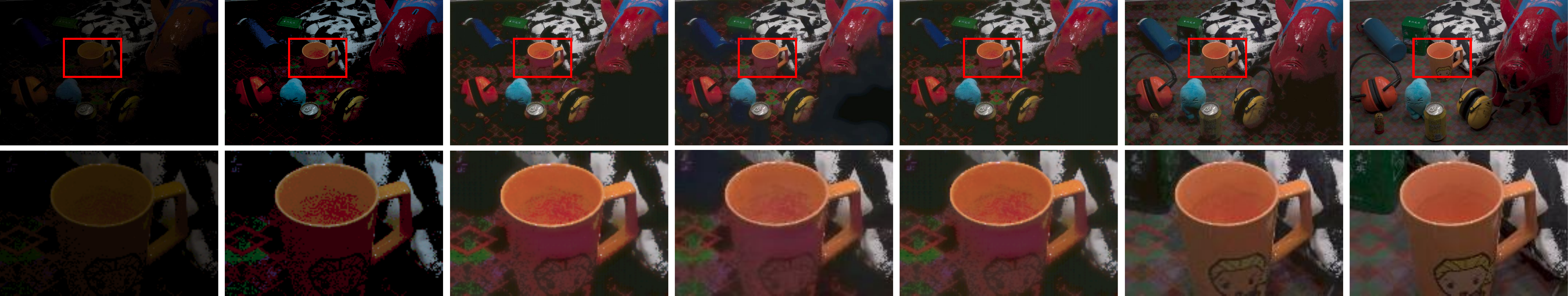}\\
    
    \includegraphics[width=\textwidth]{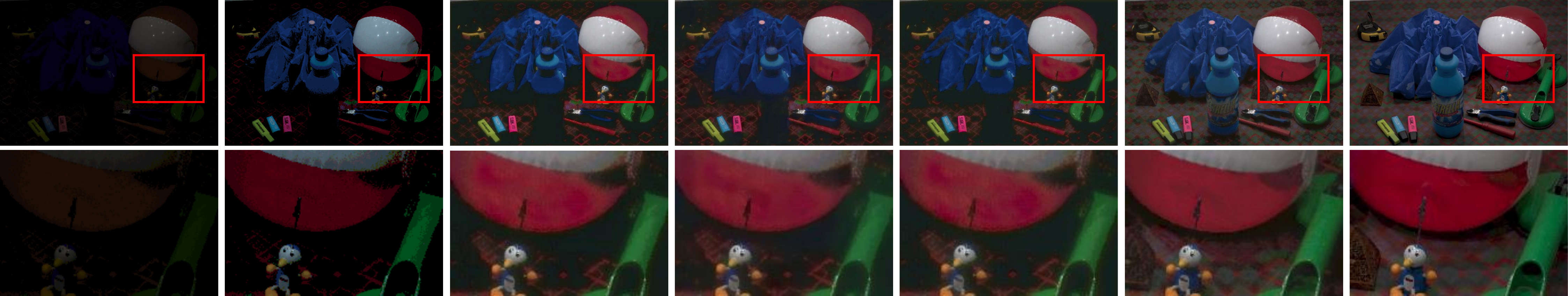}\\
    
    \includegraphics[width=\textwidth]{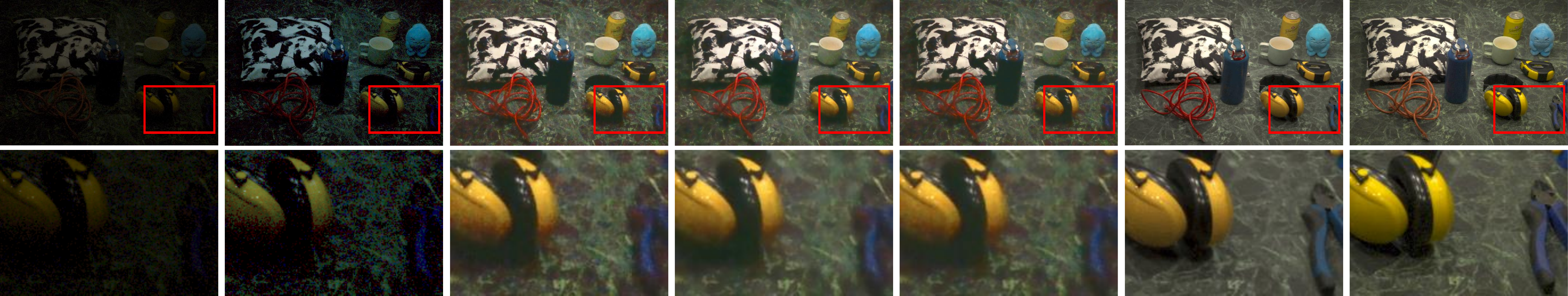}\\
    \makebox[0.138\linewidth]{\small Input}
    \makebox[0.138\linewidth]{\small SCI~\cite{SCI}}
    \makebox[0.138\linewidth]{\small GT-Mean~\cite{liao2025gt}}
    \makebox[0.138\linewidth]{\small PromptNorm~\cite{promptnorm}}
    \makebox[0.138\linewidth]{\small Retinexformer~\cite{Retinexformer}}
    \makebox[0.138\linewidth]{\small Ours}
    \makebox[0.138\linewidth]{\small Ground Truth}
    \vspace{-2mm}
    \caption{Visual comparison on MILL-s. From left to right: input, SCI~\cite{SCI}, GT-Mean~\cite{liao2025gt}, PromptNorm~\cite{promptnorm}, Retinexformer~\cite{Retinexformer}, Ours, and ground truth. We show three examples with zoomed-in regions below each. First two images: DSLR; last image: smartphone.}
    \vspace{-4mm}
    \label{fig:qualitative}
\end{figure*}

\begin{table}[t!]\caption{Quantitative comparisons on MILL-f.}
\vspace{-2mm}
\small
\setlength{\tabcolsep}{4pt}
\centering
\begin{tabular}{llccccc}
\toprule
& & PSNR$_L$ & PSNR$_C$ & SSIM & $\Delta E_{76}$ \\ \midrule

\multirow{4}{*}{\rotatebox{90}{DSLR}} & Retinexformer \cite{Retinexformer}&  27.47 & 25.41 & 0.895  & 8.27 \\
&S-Retinexformer & 28.45 & 26.31 & 0.905 & 7.48 \\
& I-Retinexformer &  36.36 & 33.09 & 0.924& 4.25 \\ 
& Ours & \cellcolor{tabfirst}37.55 & \cellcolor{tabfirst}34.05 & \cellcolor{tabfirst}0.929 & \cellcolor{tabfirst}3.67 \\
\midrule
\multirow{4}{*}{\rotatebox{90}{Smarphone}} &Baseline \cite{Retinexformer} & 22.53 & 20.62 & 0.668 & 10.85 \\
&S-Retinexformer  & 21.02 & 19.27 & 0.645 & 12.95\\
&I-Retinexformer & 23.53 & 21.55 & 0.672 & 9.63 \\
&Ours & \cellcolor{tabfirst}24.45 & \cellcolor{tabfirst}22.37 & \cellcolor{tabfirst}0.682 & \cellcolor{tabfirst}8.53  \\
\bottomrule
\vspace{-9mm}
\end{tabular}%
\label{tab:ablation}
\end{table}

\begin{figure}[t!]
    \includegraphics[width=\linewidth]{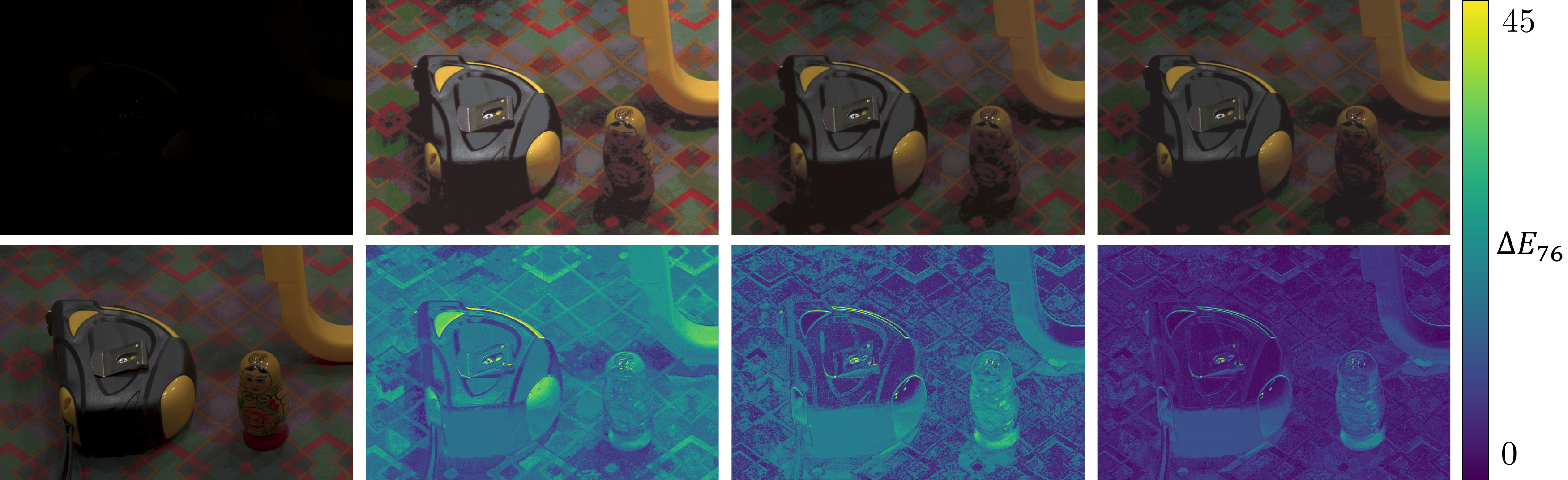}
    
    \makebox[0.22\linewidth]{\footnotesize \hspace{-2mm} Input/GT}
    \makebox[0.22\linewidth]{\footnotesize S-Retinexformer}
    \makebox[0.22\linewidth]{\footnotesize \hspace{1mm} I-Retinexformer}
    \makebox[0.22\linewidth]{\footnotesize \hspace{1mm} Ours}
    \vspace{-2mm}
    \caption{Ablation Study for the different components of our loss term. On the second row, we show the $\Delta E_{76}$ error maps.}
    \vspace{-6mm}
    \label{fig:ablation}
\end{figure}

\begin{figure*}[t!]
    \centering
    \includegraphics[width=\textwidth]{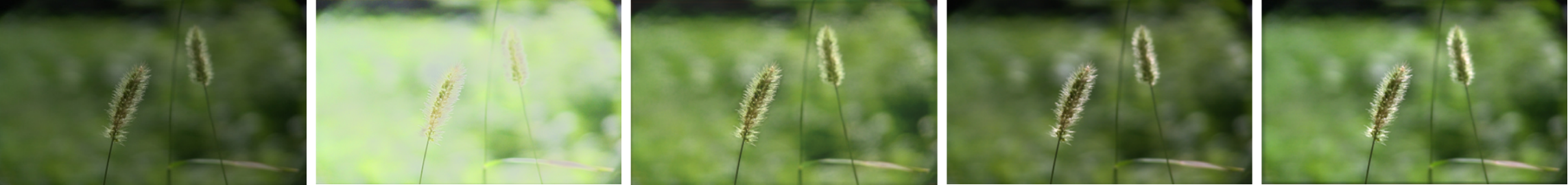}
    \\

    \includegraphics[width=\textwidth]{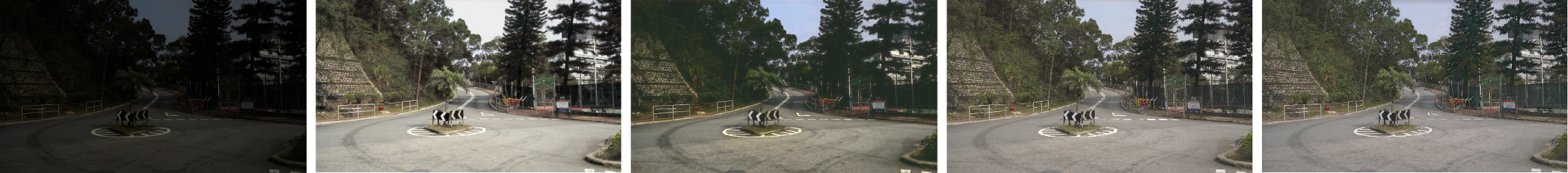}
    \\

    \makebox[0.195\linewidth]{\small Input}
    \makebox[0.195\linewidth]{\small Retinexformer (LoLv1)}
    \makebox[0.195\linewidth]{\small I-Retinexformer}
    \makebox[0.195\linewidth]{\small S-Retinexformer}
    \makebox[0.195\linewidth]{\small Ours}
    \vspace{-2mm}
    \caption{Outdoor examples from the DICM~\cite{DICM} (first row) and SICE~\cite{SICE} (second row) of Retinexformer trained on LoLv1, our baseline with the two proposed additional loss terms independently, and our final approach.}
    \vspace{-4mm}
    \label{fig:outdoor}
\end{figure*}

Table~\ref{tab:levels} presents $\Delta E_{76}$ and PSNR$_L$ results on the DSLR split of MILL-s across different brightness levels. Several interesting patterns emerge. First, certain methods, such as RUAS and FourLLIE, exhibit degraded performance as brightness increases, suggesting specialization for extremely low-light conditions at the expense of failing at correcting images at higher intensity levels. Conversely, recent methods demonstrate greater consistency across intensity levels. The first row shows input image quality: while all methods successfully enhance severely underexposed images (lower levels), most fail to improve moderately underexposed images (higher levels), indicating that robustness across varying intensities remains an open challenge. Notably, our modifications to Retinexformer yield improvements across all intensity levels (except Level 1, where PromptNorm achieves the best performance with 40 times more parameters).

Table~\ref{tab:avg} reports performance averaged across all intensity levels for both the DSLR and smartphone splits of MILL-s, using three full-reference metrics: PSNR$_L$, PSNR$_C$ (on RGB), and SSIM; one perceptual full-reference metric: LPIPS~\cite{lpips}; two color similarity metrics: $\Delta E_{76}$ and MS-SWD~\cite{ms-swd}; and two non-reference metrics: NIQE~\cite{niqe}, and Brisque~\cite{brisque}. Surprisingly, several methods fail to improve upon the input images on average. As demonstrated in Table~\ref{tab:levels}, these methods enhance extremely low-light images but degrade moderately underexposed images, resulting in net negative impact. This observation highlights the difficulty of achieving robust LLIE across variable intensity levels. Retinexformer achieves the best performance, followed closely by PromptNorm and LLFormer. This indicates that Restormer-based architectures perform well for this task. We therefore select Retinexformer as our baseline due to its superior performance and parameter efficiency. Our proposed method further improves upon Retinexformer. The smartphone split proves more challenging across all methods due to inferior sensor quality.

\subsection{Qualitative Results}
Figure~\ref{fig:qualitative} presents a qualitative comparison of our method against state-of-the-art approaches. We show three representative examples with zoomed-in regions to highlight enhancement differences. The first two examples are from the DSLR split, while the third is from the smartphone split. Across all cases, our method produces substantially better enhancements. In the first example, SCI, GT-Mean, and Retinexformer present noise and color artifacts in the background next to the orange mug and within the mug interior. PromptNorm reduces noise but fails to recover the mug's texture details. In contrast, our method produces outputs closer to ground truth with sharper object boundaries and effective noise reduction. In the second example, all competing methods generate washed-out colors with unnatural saturation and color artifacts, particularly visible in the shadow cast by the inflatable ball and the green object. In the final example, competing methods produce noisy outputs, while our method's enhanced image has higher quality with better-preserved background texture.

\subsection{FullHD Experiments and Ablation}
While the previous analysis was conducted on MILL-s due to computational constraints of older methods, we now evaluate our modifications against the best-performing baseline, Retinexformer, including an ablation study on the Full-HD MILL-f dataset. This enables assessment of our improvements without image downsampling and provides more detailed analysis of our proposed loss components. We compare Retinexformer with variants incorporating our intensity prediction loss (I-Retinexformer) and scene content loss (S-Retinexformer) independently alongside the reconstruction loss, as well as our complete method combining both losses. Table~\ref{tab:ablation} reports PSNR$_L$, PSNR$_C$, SSIM, and $\Delta E_{76}$ metrics.

Our proposed modifications outperform the baseline model across all metrics. On the DSLR split, we observe improvements of approximately 10 dB in PSNR$_L$ and PSNR$_C$, 0.03 in SSIM, and 5 in $\Delta E{76}$. The smartphone split exhibits smaller but consistent gains due to sensor limitations; nevertheless, our method maintains a clear performance advantage over the baseline. Notably, the intensity prediction loss yields larger improvements than the scene content loss when applied independently. However, combining both losses delivers the strongest performance, as effective feature disentanglement requires their joint optimization. Given the performance gap with respect to Retinexformer, we applied a global intensity adjustment to the baseline model; however, the results changed minimally, confirming that the improvements stem from feature disentanglement rather than brightness correction.

Figure~\ref{fig:ablation} shows one example of the MILL-f dataset comparing individual and combined loss terms, with corresponding $\Delta E_{76}$ error maps displayed below each output. The combined use of both loss terms achieves better performance. The error maps reveal complementary behavior: the scene content loss alone produces spatially uniform $\Delta E_{76}$ values across the image, while the intensity prediction loss concentrates errors in specific regions. Combining both loss terms reduces $\Delta E_{76}$ values both globally and locally, yielding the best overall results. This demonstrates that proper feature disentanglement is only achieved through the joint application of both loss terms.

\subsection{Outdoor Images}
We evaluate our method on underexposed outdoor images from the DICM~\cite{DICM} and SICE~\cite{SICE} datasets. Figure~\ref{fig:outdoor} presents results comparing Retinexformer trained on LoLv1 with models trained on our dataset using each of our loss terms individually and our complete approach.

In the first example, Retinexformer overexposes the scene due to the moderately low-light input, an expected limitation since LoLv1 lacks images captured at varying intensity levels. In contrast, the intensity prediction loss produces accurate exposure, while the scene content loss enhances fine details in the plant. The combination of both losses yields optimal results, balancing exposure and detail enhancement. In the second example, while Retinexformer enhances overall brightness, it oversaturates the sky region due to its higher input intensity. This highlights the fundamental limitation of LoLv1 and similar datasets that contain limited diversity and only a single fixed low-light intensity level. Our multi-level dataset mitigates this issue by learning robust LLIE across different intensity levels.

\subsection{Feature Disentanglement Analysis}

\begin{figure}
    \centering
    \includegraphics[width=.93\linewidth]{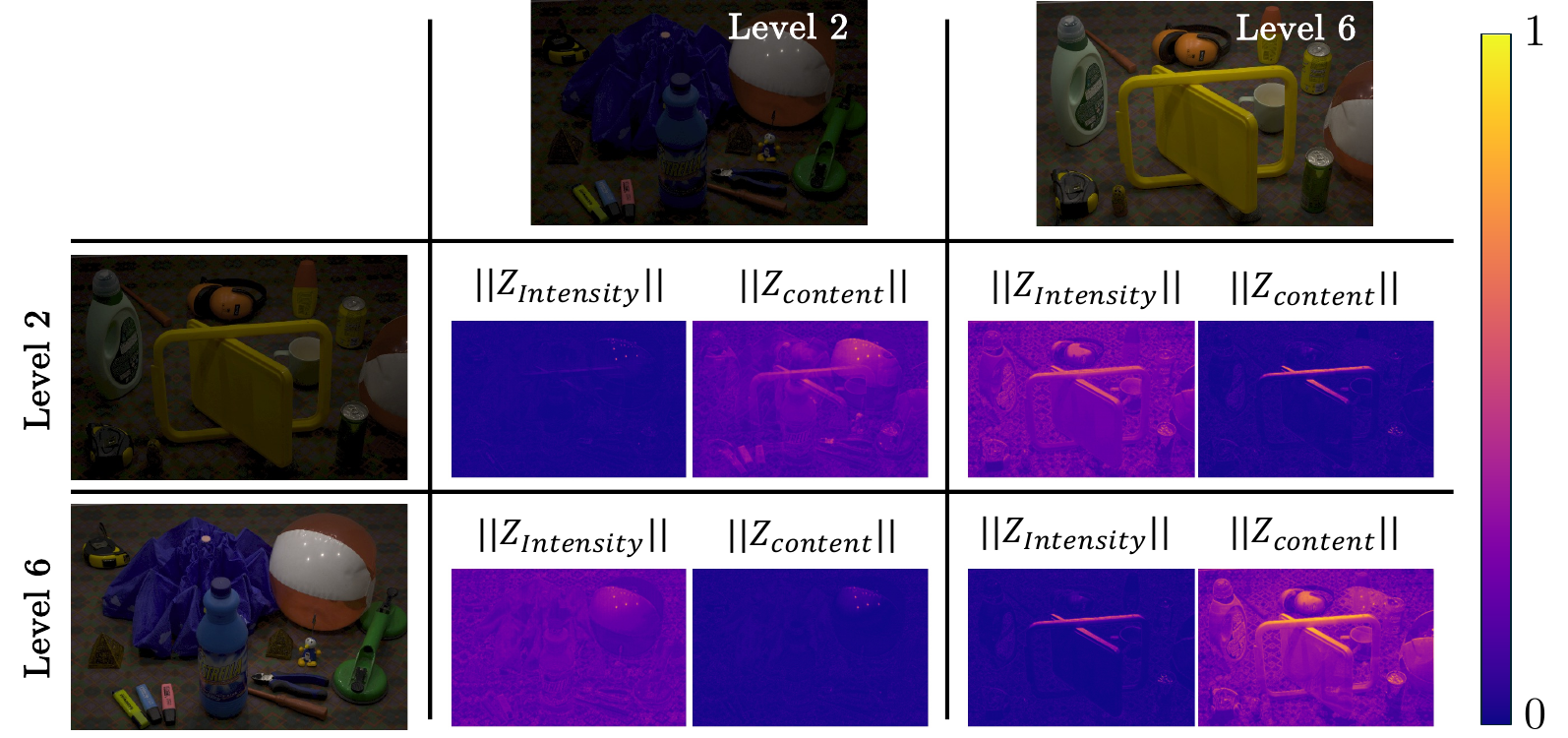}
    \caption{Feature-difference magnitudes across same-scene and different-scene pairs at varying intensity levels.}
    \vspace{-4mm}
    \label{fig:features}
\end{figure}

Fig~\ref{fig:features} shows feature‑difference magnitudes across same‑scene and different‑scene pairs at varying intensity levels. As expected, $||Z_{content}||$ is large when the scenes differ (top‑left and bottom‑right cases of the feature maps). Conversely, intensity features vary primarily with changes in lighting, as evidenced by the high $||Z_{Intensity}||$ values in the bottom‑left and top‑right cases. This validates our disentanglement claim beyond ablation studies and results.

\section{Conclusion}
We introduced the MILL dataset, which captures images under systematically varied intensity levels with all camera parameters fixed. The highest-intensity image serves as ground truth, while the remaining images serve as low-light inputs. We analyzed how current LLIE methods perform under different input intensities, revealing that performance varies significantly across methods and that robust LLIE across varying intensities remains challenging. We also propose two new loss terms that disentangle latent features into illumination intensity and scene content components, yielding substantial gains across all MILL splits.

\section*{Acknowledgements}
This work was supported by Grants PID2021-128178OB-I00 and PID2024-162555OB-I00 funded by MCIN/AEI/10.13039/ 501100011033 and by ERDF "A way of making Europe", and by the Generalitat de Catalunya CERCA Program. DSL also acknowledges the FPI grant from Spanish Ministry of Science and Innovation (PRE2022-101525). JVC also acknowledges the 2025 Leonardo Grant for Scientific Research and Cultural Creation from the BBVA Foundation. The BBVA Foundation accepts no responsibility for the opinions, statements and contents included in the project and/or the results thereof, which are entirely the responsibility of the authors. This research was also supported by the Natural Sciences and Engineering Research Council of Canada (NSERC) and the Canada Research Chairs (CRC) program.

{
    \small
    \bibliographystyle{ieeenat_fullname}
    \bibliography{main}
}

\clearpage

\appendix

\section{Extended Motivation}
Existing Low-Light Image Enhancement (LLIE) datasets present a critical limitation: they either contain only a single severely underexposed image per scene, or they simulate brightness variations through camera parameter adjustments or post-processing operations. These constraints limit real-world applicability, where low-light conditions span a continuous range of intensities.

\begin{figure}[t!] \centering
    \includegraphics[width=\linewidth]{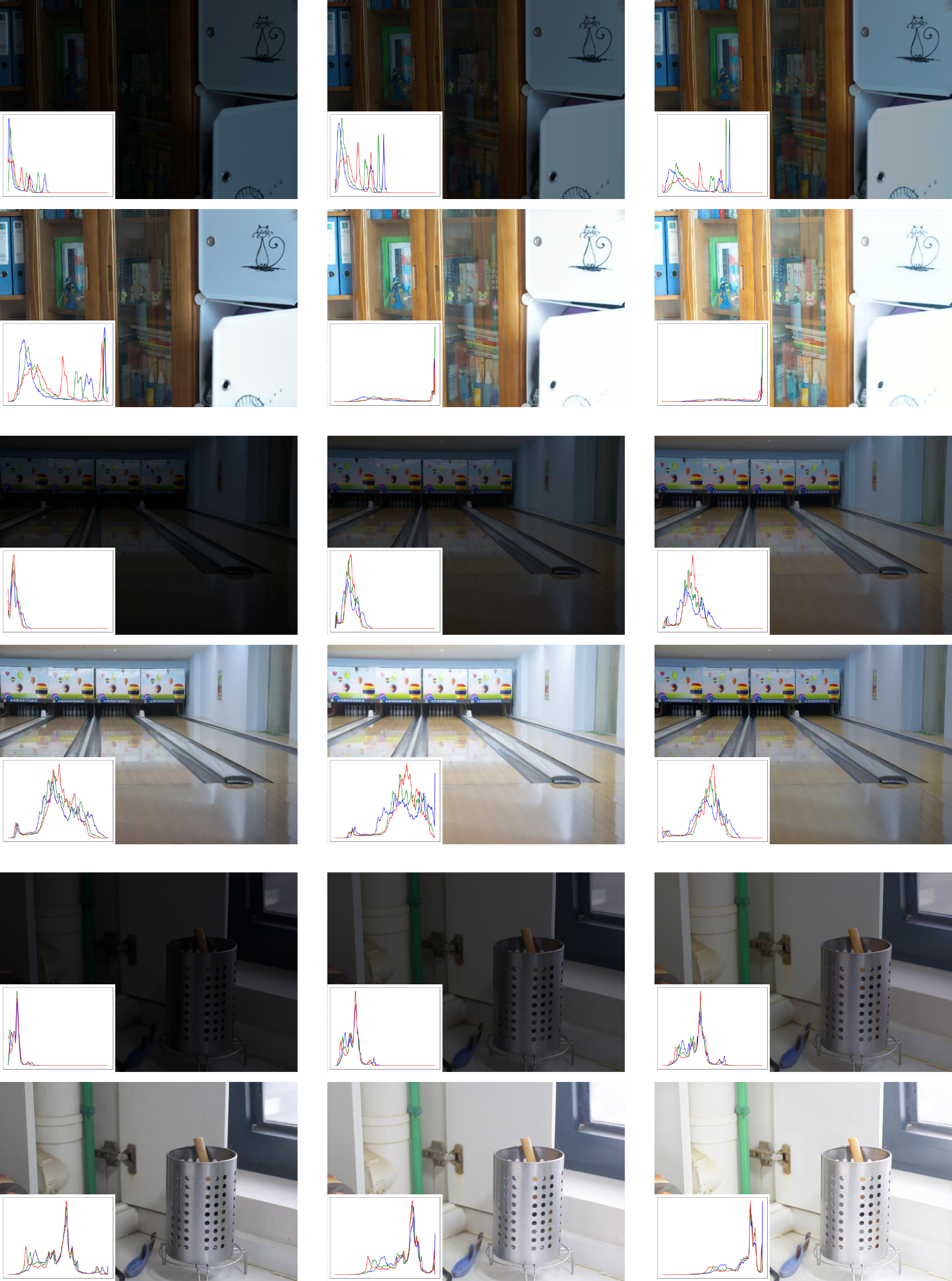}\\
    \makebox[0.325\linewidth]{\small Original}
    \makebox[0.325\linewidth]{\small 20\%}
    \makebox[0.325\linewidth]{\small 50\%}\\
    \vspace{-2mm}
    \caption{Impact of brightness variation on LLIE model performance. Blending input images with ground truth at 20\% and 50\% ratios degrades CIDNet~\cite{hvi} performance.}
    \vspace{-6mm}
    \label{fig:supp_motivation}
\end{figure}

In the main submission, we demonstrate how Retinexformer~\cite{Retinexformer} and HVI-CIDNet~\cite{hvi} exhibit degraded performance on the LoLv1 dataset when input image brightness is increased. Figure~\ref{fig:supp_motivation} provides additional examples with corresponding histograms to further demonstrate how this inherent limitation of existing LLIE datasets hinders the applicability of enhancement methods in real-world scenarios. For each example, the top row displays input images with varying brightness levels: the original image alongside versions blended with the ground truth at ratios of 0.2 and 0.5. The bottom row shows the corresponding HVI-CIDNet~\cite{hvi} outputs, with RGB histograms displayed in the bottom-left corner of each image.

In the first example, we observe that higher input intensity leads to increasingly saturated outputs. The output histograms reveal severe oversaturation, particularly evident in the large white regions of the image. The second example presents a more complex behavior: oversaturation is non-monotonic, with the 20\% blend producing more saturation than the 50\% blend. This non-linear response indicates that predicting when outputs will become oversaturated is challenging and depends on the specific image content and brightness level. Finally, the third example demonstrates how the model oversaturates the bright regions while attempting to enhance darker areas.

\begin{figure*}
    \centering
    \includegraphics[width=\textwidth]{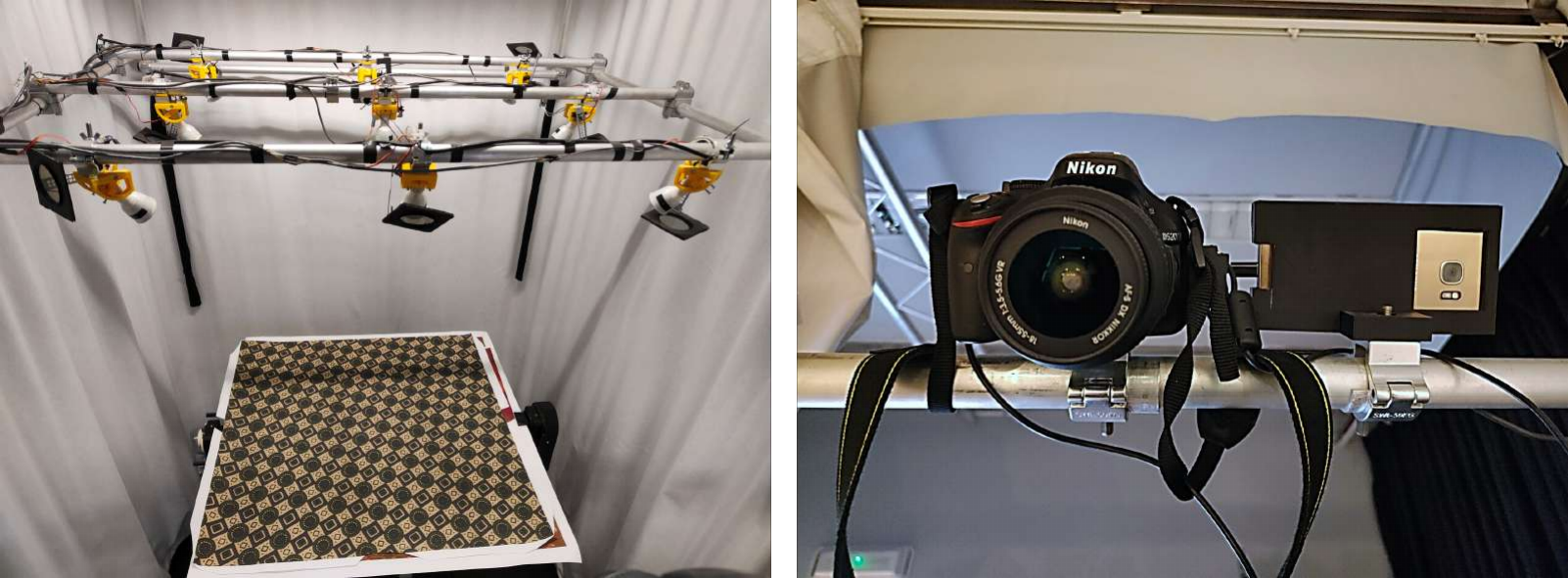}
    \vspace{-6mm}
    \caption{MILL dataset capture setup. Left: The capture platform with metallic overhead structure supporting programmable lighting arrays. Right: DSLR and smartphone cameras mounted on the structure and directed toward the platform.}
    \vspace{-4mm}
    \label{fig:supp_setup}
\end{figure*}

\begin{figure}
    \centering
    \includegraphics[width=\linewidth]{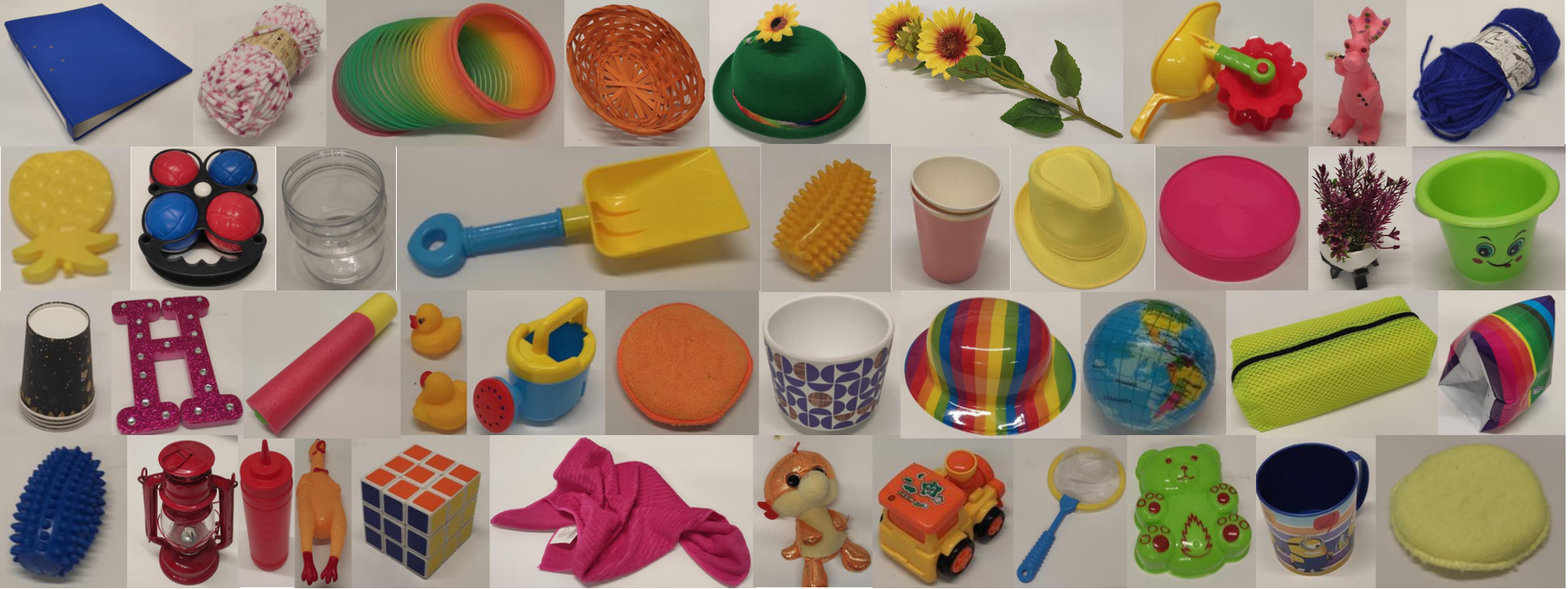}\\
    \small (a) Training objects\\
    \vspace{2mm}
    \includegraphics[width=\linewidth]{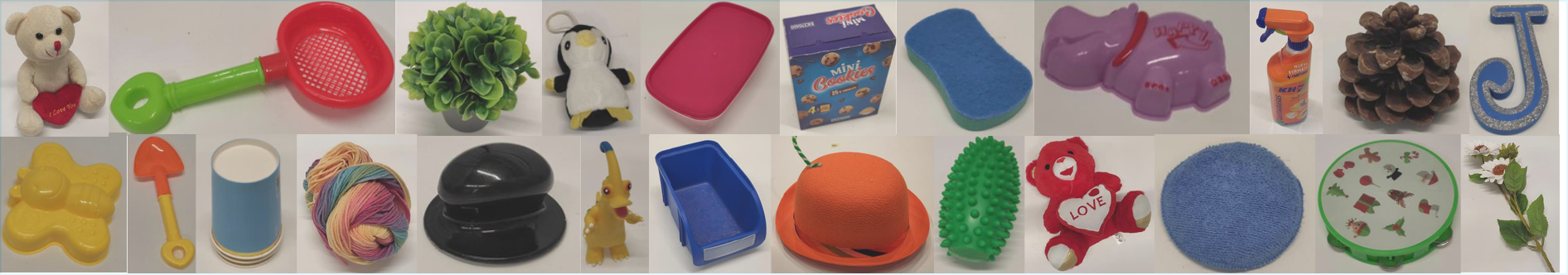}\\
    \small (b) Validation objects\\
    \vspace{2mm}
    \includegraphics[width=\linewidth]{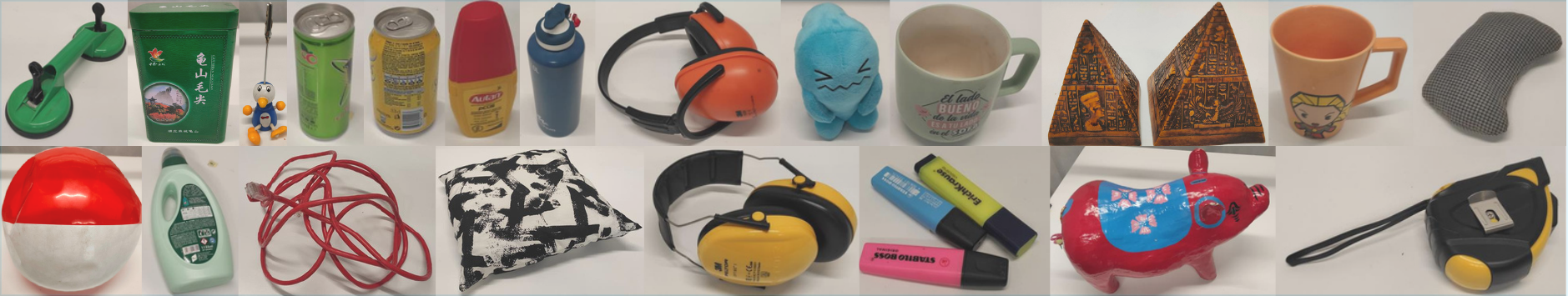}\\
    \small (c) Test objects\\
    \vspace{-1mm}
    \caption{Objects used to construct the MILL dataset.}
    \vspace{-4mm}
    \label{fig:supp_objects}
\end{figure}

\begin{figure}
    \centering
    \includegraphics[width=\linewidth]{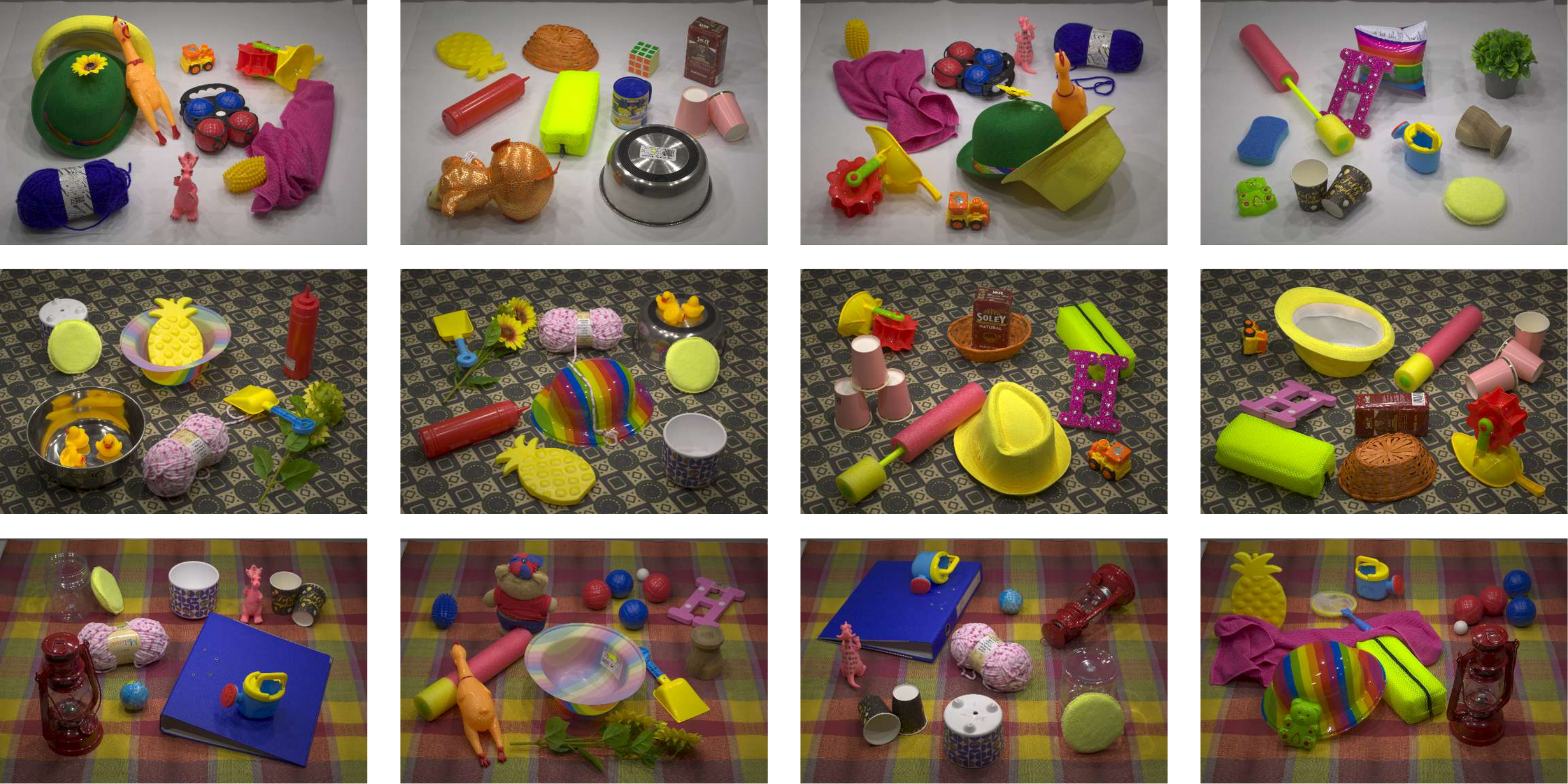}\\
    \small (a) Training scenes\\
    \vspace{2mm}
    \includegraphics[width=\linewidth]{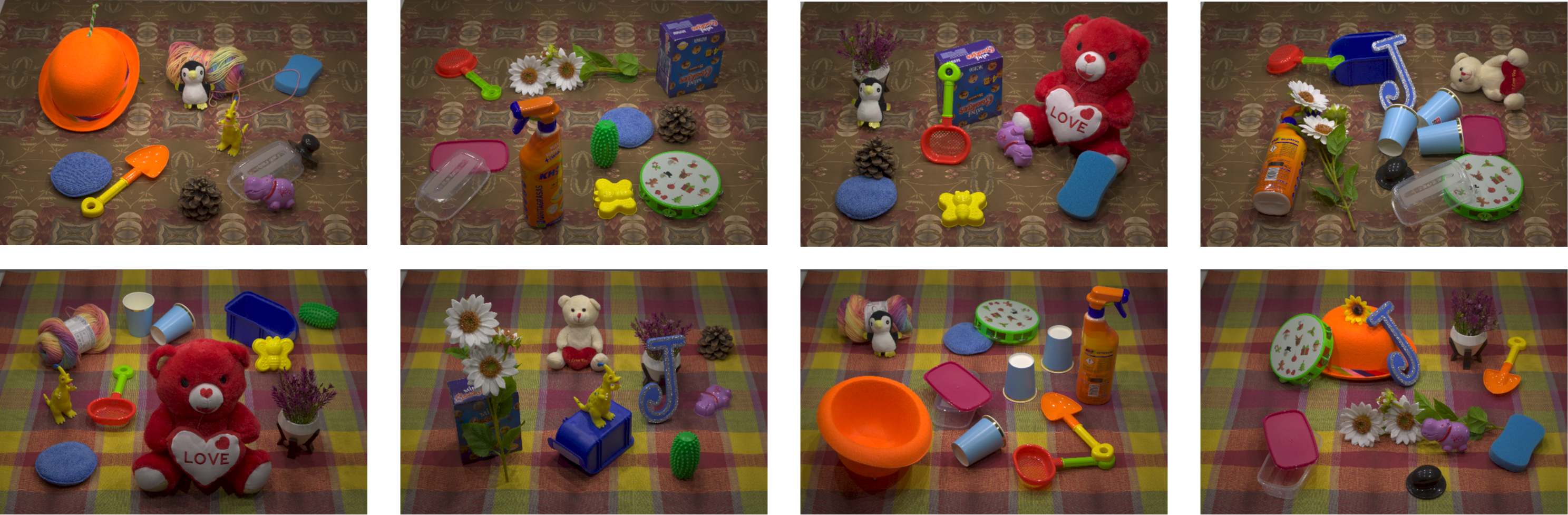}\\
    \small (b) Validation scenes\\
    \vspace{2mm}
    \includegraphics[width=\linewidth]{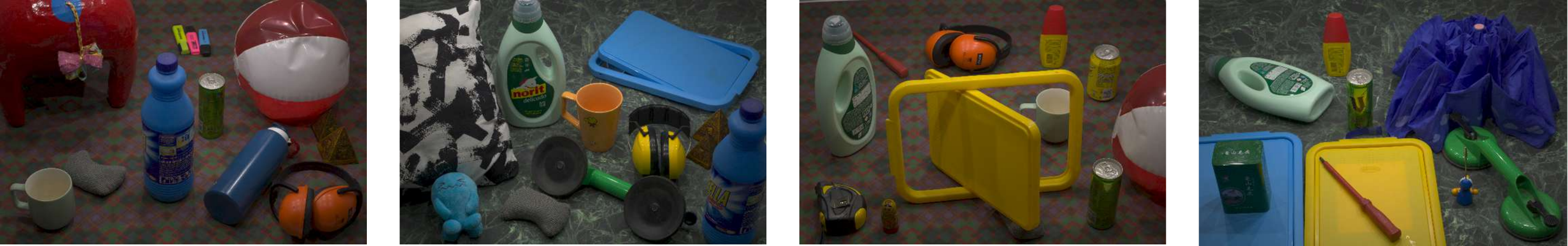}\\
    \small (c) Test scenes\\
    \vspace{-2mm}
    \caption{Representative scenes from the MILL dataset, separated by splits.}
    
    \label{fig:supp_dataset}
\end{figure}

LLIE research has mainly focused on architectural improvements to enhance performance on existing benchmarks. While these methods typically perform well on standard datasets, the datasets themselves have received considerably less attention. Through this analysis, we demonstrate that achieving robust LLIE methods requires equal focus on dataset design. This motivation drives our introduction of the Multi-Illuminant Low-Light (MILL) dataset. MILL enables systematic evaluation of current methods across different brightness intensities and provides a foundation for training more robust models, thereby advancing both LLIE research and the practical applicability of enhancement methods.

\section{Dataset Setup Details}
Acquiring images under a range of light intensities requires a controlled environment. We capture the MILL dataset in a room without windows or external light sources to eliminate uncontrolled illumination. As shown in Figure~\ref{fig:supp_setup}, our setup consists of a platform with a fixed metallic structure supporting nine programmable lights, a DSLR camera, and a smartphone. The platform accepts interchangeable floor backgrounds and allows object placement under controlled conditions.

Figure~\ref{fig:supp_objects} displays the objects used in MILL, organized by training, validation, and test splits. As mentioned in the main paper, we ensure that i) no objects appear in multiple splits, and ii) backgrounds are not shared between train/validation and test; thus maintaining strict separation to evaluate proper performance. Figure~\ref{fig:supp_dataset} shows representative images from each split, demonstrating the diversity achieved through varied object positions, orientations, and spatial arrangements.

\section{Additional Quantitative Results}
Tables~\ref{tab:levels} and~\ref{tab:level10} report the PSNR$_L$, SSIM, LPIPS, and $\Delta$E$_{76}$ metrics across all 10 illumination levels of the DSLR split in MILL-s.

\section{New Loss Terms in Other Baselines}
Table~\ref{tab:supp_ablation} presents additional results for the experiment in the camera split of the MILL-f dataset, including an ablation study. We report PSNR$_L$, PSNR$_C$, SSIM, and $\Delta$E$_{76}$ for a baseline LLIE model, with our scene content loss (S) and intensity prediction loss (I) added independently, as well as the combined loss (SI). We include DarkIR~\cite{darkir} and HVI-CIDNet~\cite{hvi} as representative state-of-the-art LLIE architectures. Each loss term independently improves performance; furthermore, combining both terms yields the best results, as effective disentanglement cannot be achieved using either term alone.

\begin{figure*}[t]
    \centering
    \includegraphics[width=\linewidth]{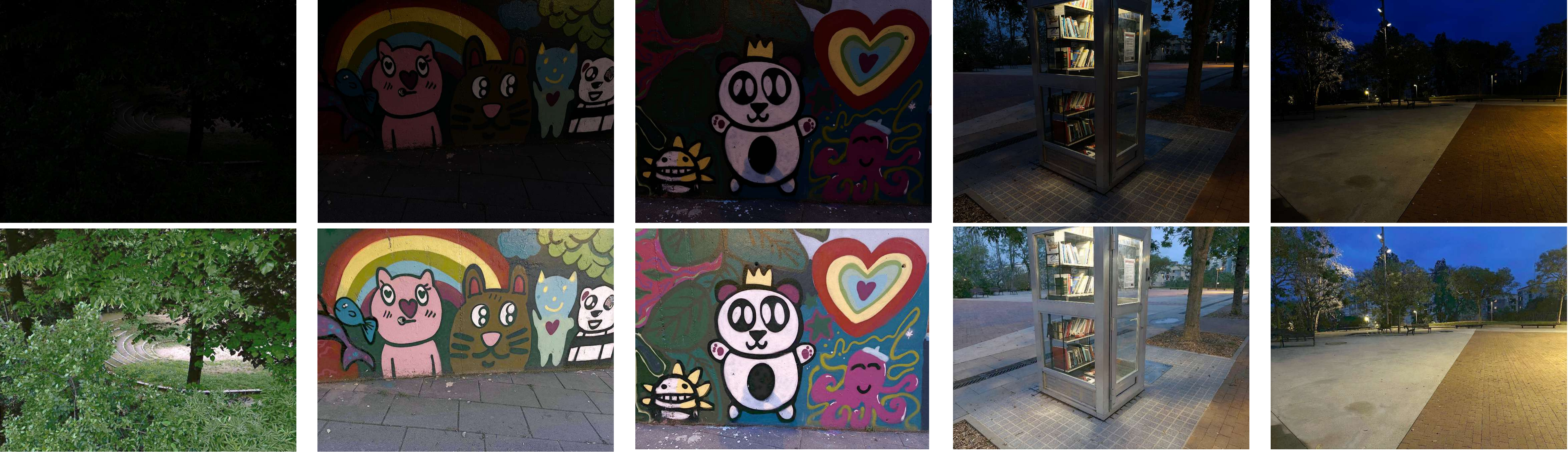}\\
    \caption{Enhancement results of our approach on outdoor low-light images captured in the wild.}
    \label{fig:supp_outdoors_ours}
\end{figure*}

\begin{figure}[t!]
    \centering
    \includegraphics[width=\linewidth]{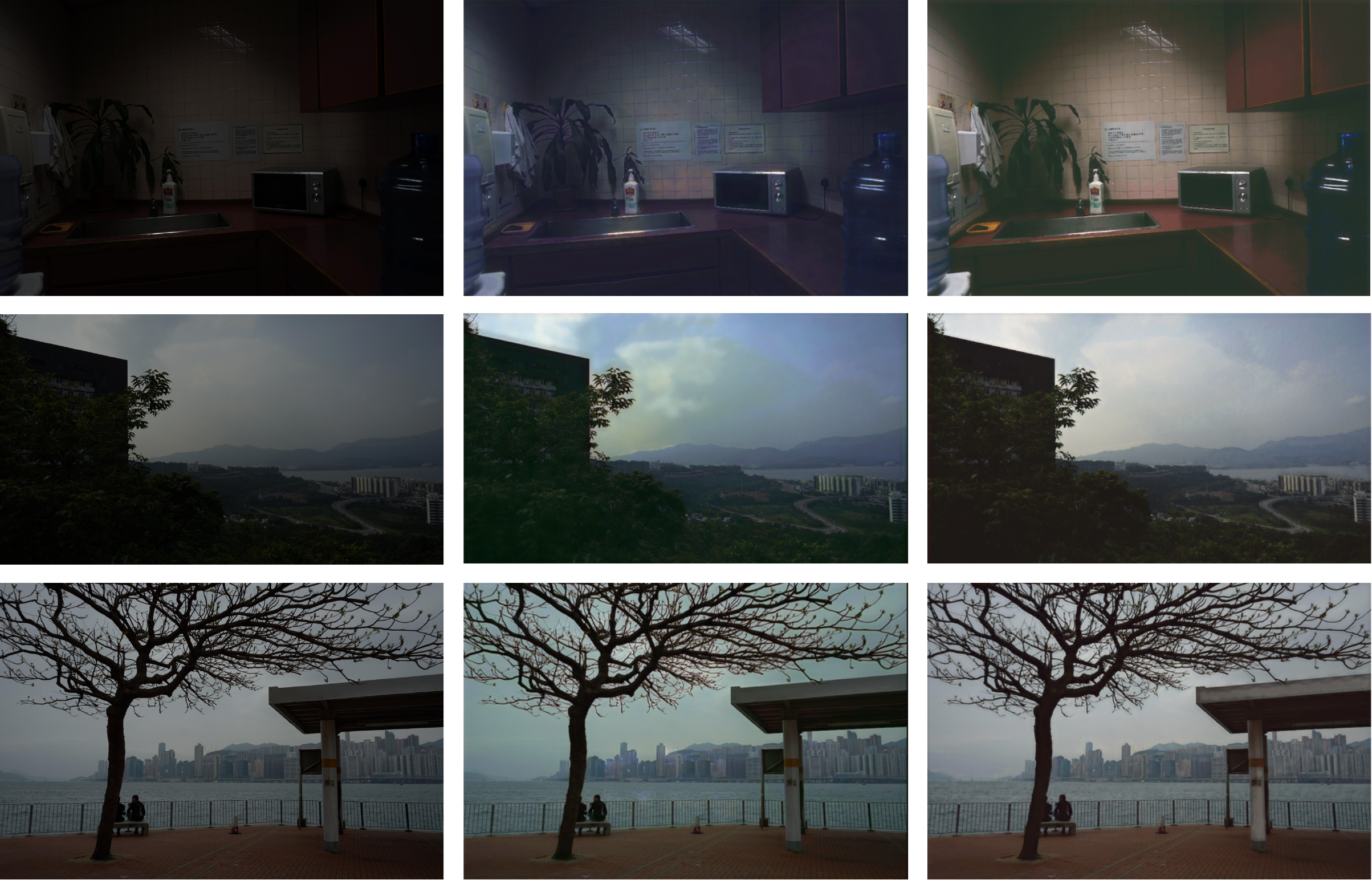}\\
    \makebox[0.32\linewidth]{\small Input}
    \makebox[0.32\linewidth]{\small Retinexformer~\cite{Retinexformer}}
    \makebox[0.32\linewidth]{\small Ours}
    \small (a) SICE~\cite{SICE}\\
    \vspace{2mm}
    \includegraphics[width=\linewidth]{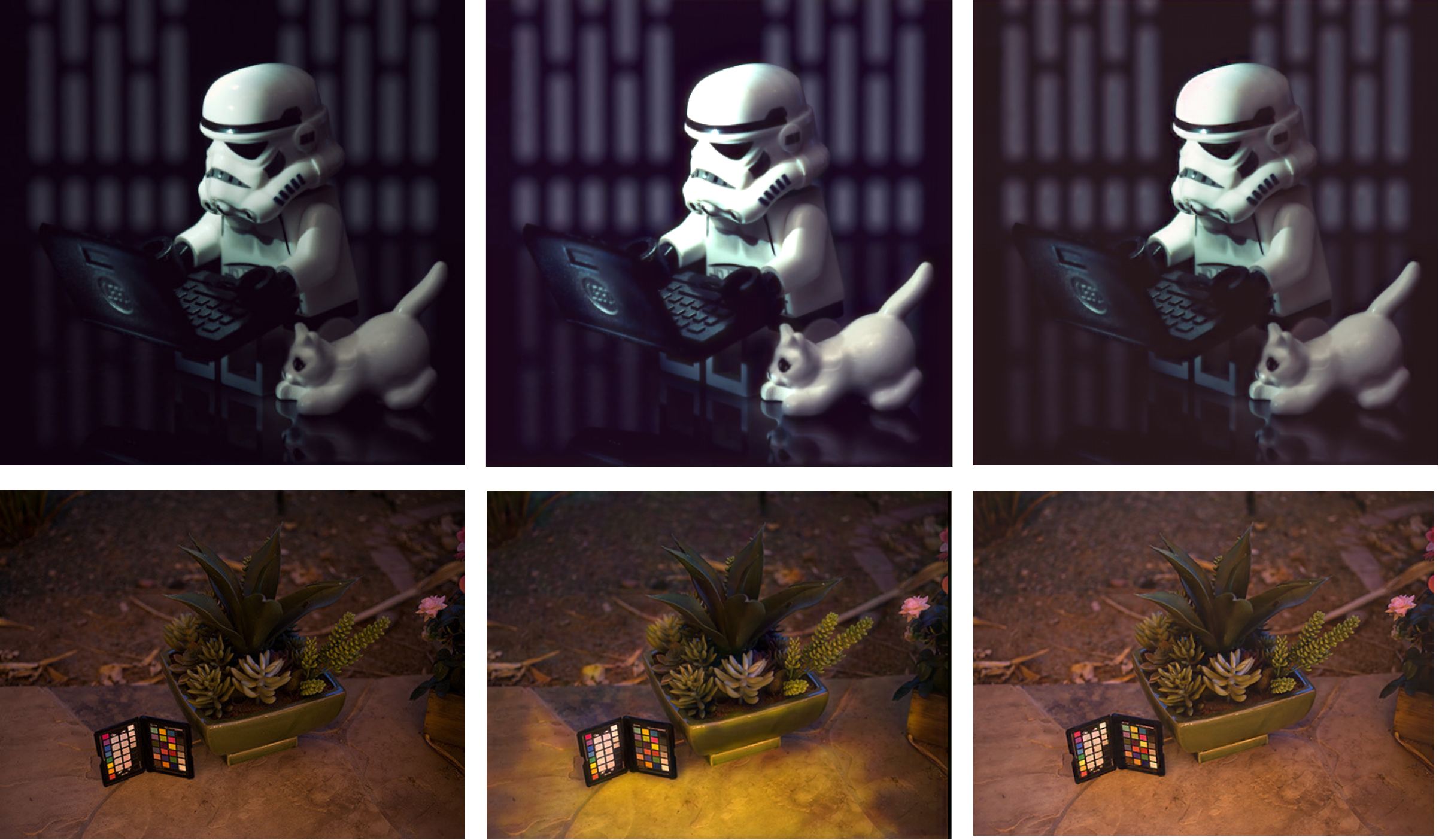}\\
    \makebox[0.32\linewidth]{\small Input}
    \makebox[0.32\linewidth]{\small Retinexformer~\cite{Retinexformer}}
    \makebox[0.32\linewidth]{\small Ours}
    \small (b) LIME~\cite{LIME}\\
    \vspace{-1mm}
    \caption{Outdoor images from SICE and LIME datasets. We display results from Retinexformer and our approach trained on MILL.}
    \label{fig:supp_outdoors_dataset}
\end{figure}

Retinexformer achieves the best performance on MILL-f and shows the greatest improvement when augmented with our loss terms (see red increments in Table~\ref{tab:supp_ablation}). We hypothesize that this is because both DarkIR and HVI-CIDNet incorporate highly specialized components for LLIE, such as dedicated color space transformations and task-specific losses (edge losses, guiding losses, or losses in alternative color spaces). In contrast, Retinexformer serves as a powerful general-purpose baseline without such domain-specific design choices, making it more receptive to our proposed loss terms.

\section{Additional Qualitative Results}
\begin{figure*}
    \centering
    \includegraphics[width=\linewidth]{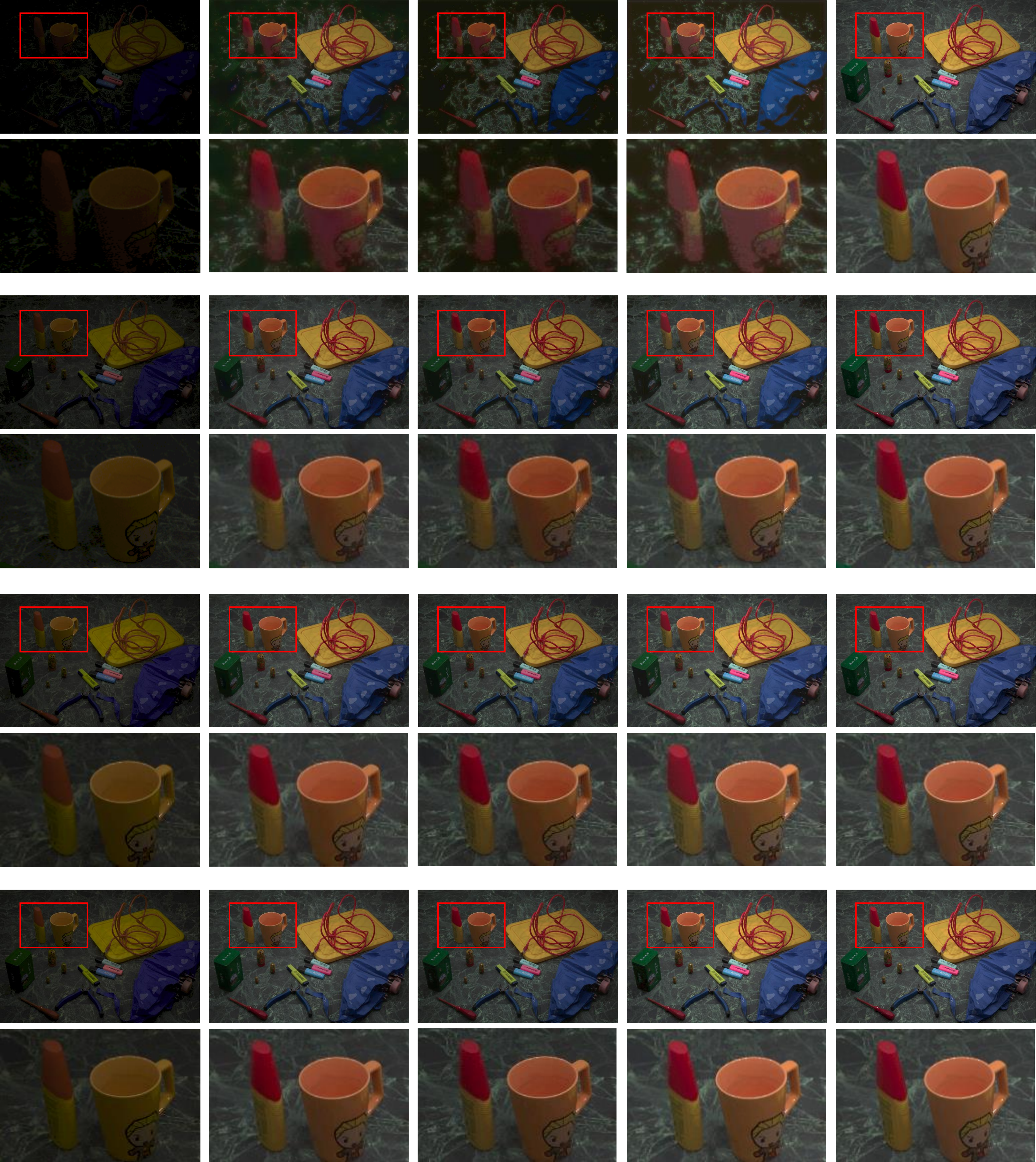}
    \makebox[0.195\linewidth]{Input}
    \makebox[0.195\linewidth]{PromptNorm~\cite{promptnorm}}
    \makebox[0.195\linewidth]{Retinexformer~\cite{Retinexformer}}
    \makebox[0.195\linewidth]{Ours}
    \makebox[0.195\linewidth]{Ground Truth}
    \caption{First four levels of a scene of MILL with results from PromptNorm~\cite{promptnorm}, Retinexformer~\cite{Retinexformer} and our approach.}
    \label{fig:supp_qualitative_levels}
\end{figure*}

Figure~\ref{fig:supp_qualitative_levels} shows a MILL-s scene captured at the first four illumination levels, with enhancement results from PromptNorm~\cite{promptnorm}, Retinexformer~\cite{Retinexformer}, and our approach. For each image, we provide a zoomed-in region displayed below. Our method produces results closer to the ground truth, as evidenced by the accurate color reproduction in the red and yellow pen and the orange mug. Note that at Level 1 (the most challenging condition), all methods produce outputs that deviate considerably from the ground truth. However, as the illumination level increases, all methods improve, benefiting from the additional information preserved in the input images.

\begin{figure*}
    \centering
    \includegraphics[width=\linewidth]{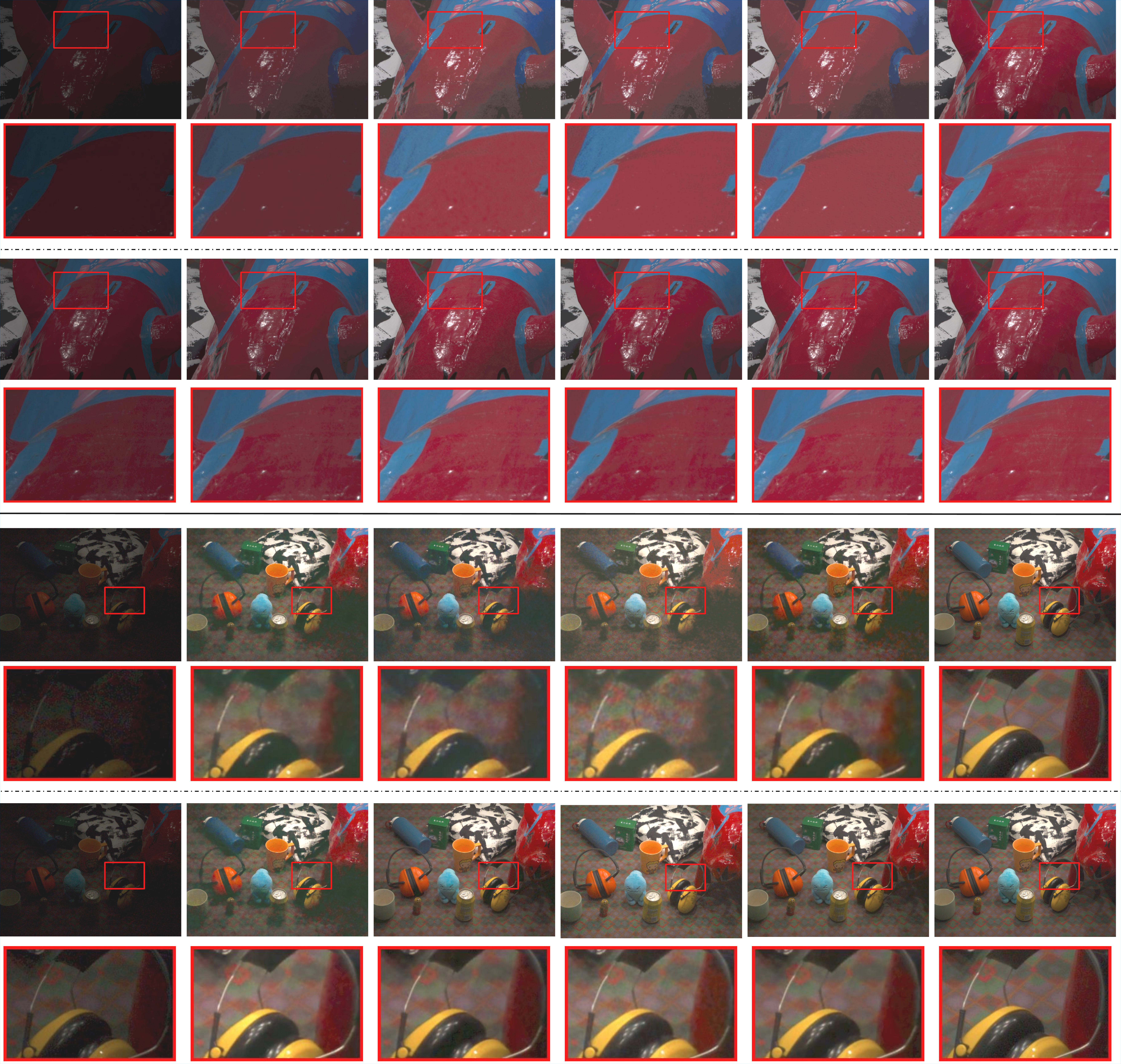}
    \makebox[0.162\linewidth]{Input}
    \makebox[0.162\linewidth]{Retinexformer~\cite{Retinexformer}}
    \makebox[0.162\linewidth]{I-Retinexformer}
    \makebox[0.162\linewidth]{S-Retinexformer}
    \makebox[0.162\linewidth]{Ours}
    \makebox[0.162\linewidth]{Ground Truth}
    \caption{Qualitative comparison on MILL-f (top) and MILL-s (bottom) at Level 1 and Level 4. We compare Retinexformer~\cite{Retinexformer} with our loss terms applied independently (S and I) and combined (Ours).}    \label{fig:supp_ablation}
\end{figure*}

Figure~\ref{fig:supp_ablation} presents results on one MILL-f scene and one MILL-s scene, each evaluated at two illumination levels (Level 1 and Level 4). We compare Retinexformer~\cite{Retinexformer}, our loss terms applied independently, and our combined approach. Zoomed-in crops of visually salient regions are displayed below each result. Our combined approach achieves superior quality at Level 1 for both examples, while at Level 4 it produces sharper details and more accurate color reproduction compared to the ground truth.

Figure~\ref{fig:supp_outdoors_dataset} presents qualitative comparisons between Retinexformer~\cite{Retinexformer} and our approach on outdoor images from the SICE~\cite{SICE} and LIME~\cite{LIME} datasets. Both methods are trained on MILL-s. Figure~\ref{fig:supp_outdoors_ours} demonstrates the generalization capability of our method on in-the-wild outdoor images.

\begin{table*}[t!] 
\caption{Levels 1 to 9 of the DSLR MILL-s dataset.}
\setlength{\tabcolsep}{2.6pt}
\vspace{-2mm}
\centering
\small
\begin{tabular}{lcccccccccccc}
\toprule
& \multicolumn{4}{c}{Level 1} & \multicolumn{4}{c}{Level 2} & \multicolumn{4}{c}{Level 3} \\
\cmidrule{2-13}
\multicolumn{1}{c}{DSLR} & PSNR$_L$ $\uparrow$ & SSIM $\uparrow$ & LPIPS $\downarrow$ & $\Delta$E$_{76}$ $\downarrow$
& PSNR$_L$ $\uparrow$ & SSIM $\uparrow$ & LPIPS $\downarrow$ & $\Delta$E$_{76}$ $\downarrow$
& PSNR$_L$ $\uparrow$ & SSIM $\uparrow$ & LPIPS $\downarrow$ & $\Delta$E$_{76}$ $\downarrow$\\
\midrule
Unprocessed & 13.457 & 0.132 & 0.483 & 30.337 & 15.884 & 0.443 & 0.258 & 23.108 & 17.681 & 0.623 & 0.157 & 18.621 \\
\hdashline
RUAS~\cite{Ruas} & 16.635 & 0.316 & 0.457 & 25.462 & 13.068 & 0.438 & 0.394 & 36.526 & 9.861 & 0.426 & 0.399 & 45.332 \\
LLFormer~\cite{LLFormer} & 20.881 & 0.712 & 0.378 & 16.366 & 21.636 & 0.804 & 0.201 & 14.206 & 21.790 & 0.838 & 0.154 & 13.735 \\
KinD~\cite{KinD} & 16.706 & 0.571 & 0.439 & 23.884 & 17.900 & 0.647 & 0.324 & 21.865 & 19.756 & 0.738 & 0.247 & 17.615 \\
FourLLIE~\cite{wang2023fourllie} & 17.287 & 0.434 & 0.458 & 24.510 & 19.726 & 0.671 & 0.306 & 21.114 & 17.658 & 0.741 & 0.240 & 22.791 \\
SCI~\cite{SCI} & 16.020 & 0.285 & 0.460 & 24.049 & 23.577 & 0.617 & 0.304 & 16.431 & 21.423 & 0.696 & 0.243 & 17.994 \\
MirNet~\cite{mirnet} & 26.458 & 0.769 & 0.312 & 14.030 & 25.426 & 0.848 & 0.167 & 11.811 & 25.336 & 0.877 & 0.133 & 11.106 \\
Retinexformer~\cite{Retinexformer} & 25.092 & 0.742 & 0.335 & 14.147 & 25.945 & 0.838 & 0.180 & 11.969 & 26.390 & 0.881 & 0.137 & 10.449 \\
DarkIR~\cite{darkir} & 24.651 & 0.736 & 0.336 & 14.392 & 25.522 & 0.833 & 0.182 & 12.367 & 25.233 & 0.873 & 0.139 & 11.293 \\
CIDNet~\cite{hvi} & 24.080 & 0.725 & 0.340 & 14.781 & 23.800 & 0.812 & 0.195 & 13.606 & 22.488 & 0.845 & 0.153 & 13.713 \\
PromptNorm~\cite{promptnorm} & 25.886 & 0.770 & 0.310 & 13.471 & 25.973 & 0.854 & 0.164 & 11.507 & 26.065 & 0.888 & 0.128 & 10.513 \\
GT-Mean~\cite{liao2025gt} & 24.320 & 0.731 & 0.338 & 14.593 & 24.514 & 0.822 & 0.189 & 13.127 & 23.760 & 0.860 & 0.145 & 12.478 \\
Ours & 25.534 & 0.754 & 0.348 & 13.896 & 30.256 & 0.863 & 0.167 & 9.918 & 31.474 & 0.897 & 0.126 & 9.113 \\
\midrule
& \multicolumn{4}{c}{Level 4} & \multicolumn{4}{c}{Level 5} & \multicolumn{4}{c}{Level 6} \\
\cmidrule{2-13}
\multicolumn{1}{c}{DSLR} & PSNR$_L$ $\uparrow$ & SSIM $\uparrow$ & LPIPS $\downarrow$ & $\Delta$E$_{76}$ $\downarrow$
& PSNR$_L$ $\uparrow$ & SSIM $\uparrow$ & LPIPS $\downarrow$ & $\Delta$E$_{76}$ $\downarrow$
& PSNR$_L$ $\uparrow$ & SSIM $\uparrow$ & LPIPS $\downarrow$ & $\Delta$E$_{76}$ $\downarrow$\\
\midrule
Unprocessed & 19.480 & 0.742 & 0.104 & 15.024 & 21.485 & 0.829 & 0.068 & 11.897 & 23.341 & 0.880 & 0.047 & 9.649 \\
\hdashline
RUAS~\cite{Ruas} & 8.029 & 0.394 & 0.434 & 52.381 & 6.931 & 0.362 & 0.475 & 57.468 & 6.301 & 0.343 & 0.510 & 60.609 \\
LLFormer~\cite{LLFormer} & 21.918 & 0.854 & 0.134 & 13.516 & 22.061 & 0.868 & 0.123 & 13.343 & 22.158 & 0.874 & 0.118 & 13.246 \\
KinD~\cite{KinD} & 20.656 & 0.789 & 0.208 & 15.814 & 21.340 & 0.821 & 0.183 & 14.872 & 21.679 & 0.836 & 0.170 & 14.506 \\
FourLLIE~\cite{wang2023fourllie} & 15.732 & 0.731 & 0.218 & 25.400 & 14.968 & 0.730 & 0.203 & 26.642 & 14.514 & 0.725 & 0.198 & 27.660 \\
SCI~\cite{SCI} & 17.983 & 0.698 & 0.228 & 21.918 & 15.691 & 0.681 & 0.228 & 25.890 & 14.314 & 0.661 & 0.233 & 28.892 \\
MirNet~\cite{mirnet} & 24.974 & 0.885 & 0.121 & 11.295 & 24.814 & 0.895 & 0.115 & 11.394 & 24.605 & 0.899 & 0.113 & 11.551 \\
Retinexformer~\cite{Retinexformer} & 26.548 & 0.894 & 0.122 & 10.463 & 26.548 & 0.907 & 0.113 & 10.353 & 26.476 & 0.912 & 0.108 & 10.389 \\
DarkIR~\cite{darkir} & 24.809 & 0.883 & 0.123 & 11.644 & 24.742 & 0.896 & 0.113 & 11.579 & 24.238 & 0.898 & 0.110 & 11.918 \\
CIDNet~\cite{hvi} & 21.670 & 0.851 & 0.137 & 14.508 & 21.442 & 0.863 & 0.126 & 14.585 & 21.049 & 0.864 & 0.122 & 14.965 \\
PromptNorm~\cite{promptnorm} & 26.090 & 0.899 & 0.116 & 10.568 & 25.944 & 0.910 & 0.109 & 10.591 & 25.777 & 0.913 & 0.106 & 10.716 \\
GT-Mean~\cite{liao2025gt} & 22.910 & 0.866 & 0.130 & 13.252 & 22.801 & 0.879 & 0.119 & 13.226 & 22.571 & 0.882 & 0.115 & 13.383 \\
Ours & 31.360 & 0.895 & 0.114 & 9.091 & 31.517 & 0.908 & 0.107 & 9.088 & 32.094 & 0.917 & 0.101 & 9.017 \\
\midrule
& \multicolumn{4}{c}{Level 7} & \multicolumn{4}{c}{Level 8} & \multicolumn{4}{c}{Level 9} \\
\cmidrule{2-13}
\multicolumn{1}{c}{DSLR} & PSNR$_L$ $\uparrow$ & SSIM $\uparrow$ & LPIPS $\downarrow$ & $\Delta$E$_{76}$ $\downarrow$
& PSNR$_L$ $\uparrow$ & SSIM $\uparrow$ & LPIPS $\downarrow$ & $\Delta$E$_{76}$ $\downarrow$
& PSNR$_L$ $\uparrow$ & SSIM $\uparrow$ & LPIPS $\downarrow$ & $\Delta$E$_{76}$ $\downarrow$\\
\midrule
Unprocessed & 25.557 & 0.915 & 0.030 & 7.595 & 28.553 & 0.942 & 0.022 & 6.132 & 32.063 & 0.952 & 0.020 & 5.318 \\
RUAS~\cite{Ruas} & 5.859 & 0.329 & 0.540 & 62.992 & 5.507 & 0.318 & 0.577 & 65.126 & 5.478 & 0.313 & 0.583 & 65.994 \\
LLFormer~\cite{LLFormer} & 22.247 & 0.879 & 0.114 & 13.170 & 22.406 & 0.885 & 0.110 & 13.005 & 22.727 & 0.897 & 0.107 & 12.612 \\
KinD~\cite{KinD} & 21.627 & 0.842 & 0.161 & 14.493 & 21.680 & 0.851 & 0.155 & 14.658 & 21.531 & 0.856 & 0.150 & 14.879 \\
FourLLIE~\cite{wang2023fourllie} & 14.069 & 0.716 & 0.195 & 28.788 & 13.802 & 0.711 & 0.193 & 29.569 & 13.503 & 0.710 & 0.194 & 30.625 \\
SCI~\cite{SCI} & 13.244 & 0.640 & 0.239 & 31.655 & 12.291 & 0.620 & 0.248 & 34.747 & 11.711 & 0.615 & 0.254 & 36.754 \\
MirNet~\cite{mirnet} & 24.488 & 0.902 & 0.111 & 11.650 & 25.040 & 0.910 & 0.107 & 11.533 & 25.334 & 0.919 & 0.103 & 11.513 \\
Retinexformer~\cite{Retinexformer} & 26.480 & 0.916 & 0.104 & 10.414 & 27.156 & 0.925 & 0.100 & 10.392 & 27.658 & 0.934 & 0.096 & 10.335 \\
DarkIR~\cite{darkir} & 23.910 & 0.900 & 0.106 & 12.213 & 24.533 & 0.909 & 0.102 & 12.113 & 24.793 & 0.919 & 0.100 & 12.088 \\
CIDNet~\cite{hvi} & 20.829 & 0.865 & 0.119 & 15.222 & 20.565 & 0.866 & 0.119 & 15.836 & 20.890 & 0.880 & 0.115 & 15.549 \\
PromptNorm~\cite{promptnorm} & 25.661 & 0.916 & 0.105 & 10.817 & 26.300 & 0.924 & 0.101 & 10.750 & 26.721 & 0.932 & 0.098 & 10.732 \\
GT-Mean~\cite{liao2025gt} & 22.194 & 0.882 & 0.113 & 13.797 & 22.473 & 0.891 & 0.109 & 13.790 & 22.313 & 0.898 & 0.108 & 14.054 \\
Ours & 32.314 & 0.924 & 0.097 & 8.942 & 32.580 & 0.932 & 0.095 & 9.291 & 34.158 & 0.934 & 0.094 & 9.172\\
\bottomrule
\end{tabular}
\label{tab:levels}
\end{table*}

\begin{table}[t!] 
\caption{Levels 10 of the DSLR MILL-s dataset.}
\setlength{\tabcolsep}{2.6pt}
\vspace{-2mm}
\centering
\small
\begin{tabular}{lcccc}
\toprule
& \multicolumn{4}{c}{Level 10}\\
\cmidrule{2-5}
\multicolumn{1}{c}{DSLR} & PSNR$_L$ $\uparrow$ & SSIM $\uparrow$ & LPIPS $\downarrow$ & $\Delta$E$_{76}$ $\downarrow$\\
\midrule
Unprocessed & 36.642 & 0.965 & 0.014 & 3.620 \\
\hdashline
RUAS~\cite{Ruas} & 5.184 & 0.309 & 0.615 & 67.143 \\
LLFormer~\cite{LLFormer} & 22.552 & 0.892 & 0.108 & 12.898 \\
KinD~\cite{KinD} & 21.360 & 0.854 & 0.148 & 14.997 \\
FourLLIE~\cite{wang2023fourllie} & 13.353 & 0.700 & 0.196 & 30.970 \\
SCI~\cite{SCI} & 11.231 & 0.594 & 0.262 & 38.379 \\
MirNet~\cite{mirnet} & 24.960 & 0.916 & 0.105 & 11.716 \\
Retinexformer~\cite{Retinexformer} & 27.412 & 0.932 & 0.098 & 10.457 \\
DarkIR~\cite{darkir} & 24.635 & 0.917 & 0.101 & 12.155 \\
CIDNet~\cite{hvi} & 20.631 & 0.874 & 0.116 & 15.847 \\
PromptNorm~\cite{promptnorm} & 26.281 & 0.930 & 0.100 & 10.889 \\
GT-Mean~\cite{liao2025gt} & 22.877 & 0.902 & 0.105 & 13.571 \\
Ours & 32.476 & 0.937 & 0.094 & 9.170 \\
\bottomrule
\end{tabular}
\label{tab:level10}
\end{table}

\begin{table}[t!] 
\caption{Quantitative comparison on the MILL-f dataset. We compare DarkIR, CIDNet and Retinexformer with our loss terms added independently (S and I) and our combined approach. We also report the gains with respect to the corresponding baseline.}
\setlength{\tabcolsep}{2.6pt}
\vspace{-2mm}
\centering
\small
\begin{tabular}{lcccc}
\toprule
\multicolumn{1}{c}{DSLR} & PSNR$_L$ $\uparrow$ & PSNR$_C$ $\uparrow$ & SSIM $\uparrow$ & $\Delta$E$_{76}$ $\downarrow$\\
\midrule
DarkIR~\cite{darkir} & 24.92 & 21.87 & 0.879 & 11.35\\
S-DarkIR & 25.09 & 21.98 & 0.881 & 10.99 \\
I-DarkIR & 25.92 & 23.04 & 0.889 & 10.20\\
SI-DarkIR & 26.83 & 23.80 & 0.896 & 9.03 \\
\hdashline
& \color{red} (+1.91) & \color{red} (+1.93) & \color{red} (+0.02) & \color{red} (-2.32) \\
\midrule
CIDNet~\cite{hvi} & 22.58 & 20.49 & 0.857 & 13.76 \\
S-CIDNet & 23.16 & 21.36 & 0.864 & 12.71 \\
I-CIDNet & 25.87 & 23.61 & 0.880 & 11.04 \\
SI-CIDNet & 26.54 & 24.52 & 0.884 & 10.69 \\
\hdashline
& \color{red} (+3.96) & \color{red} (+4.03) & \color{red} (+0.03) & \color{red} (-3.07) \\
\midrule
Retinexformer~\cite{Retinexformer} & 27.47 & 25.41 & 0.895 & 8.27 \\
S-Retinexformer & 28.45 & 26.31 & 0.905 & 7.48 \\
I-Retinexformer &  36.36 & 33.09 & 0.924 & 4.25 \\ 
Ours & 37.55 & 34.05 & 0.929 & 3.67 \\
\hdashline
& \color{red} (+10.08) & \color{red} (+8.64) & \color{red} (+0.03) & \color{red} (-4.60) \\
\bottomrule
\end{tabular}
\label{tab:supp_ablation}
\end{table}

\clearpage

\end{document}